%% file: main.tex
\documentclass{article}

\PassOptionsToPackage{square,numbers}{natbib}
\usepackage[preprint]{neurips_2022}

\input{defs-kjun-v4a}

\usepackage[utf8]{inputenc} % allow utf-8 input
\usepackage[T1]{fontenc}    % use 8-bit T1 fonts
\usepackage{hyperref}       % hyperlinks
\usepackage{url}            % simple URL typesetting
\usepackage{booktabs}       % professional-quality tables
\usepackage{amsfonts}       % blackboard math symbols
\usepackage{nicefrac}       % compact symbols for 1/2, etc.
\usepackage{microtype}      % microtypography
\usepackage{xcolor}         % colors

\usepackage{MnSymbol} % from Rakhlin

\usepackage[shortlabels]{enumitem}
\setlist[itemize]{topsep=.5pt,itemsep=0pt,parsep=2pt,leftmargin=2em}
\setlist[enumerate]{topsep=.5pt,itemsep=0pt,parsep=2pt,leftmargin=2em}
\usepackage{algorithmic}

\usepackage{amsthm}
\newtheorem{theorem}{Theorem}
\newtheorem{proposition}{Proposition}

\newtheorem{lemma}{Lemma}

\def\one{\mathbbm{1}}
\def\BB{\mathbb{B}}
\def\clip#1{\wbar{\del{#1}}}
\def\brcT{\breve{\cT}}
\def\onec#1{\one\cbr{#1}}

\DeclareMathOperator{\anc}{\mathsf{anc}}
\DeclareMathOperator{\prev}{\mathsf{prev}}
\def\brx{\breve{x}} 

\def\la{{\langle}}
\def\ra{{\rangle}}
\def\det{\ensuremath{\text{det}}}
\def\hcB{\ensuremath{\hat\cB}} 
\def\sfm{\mathsf{m}}
\def\sft{\mathsf{t}}
\def\sfh{\mathsf{h}}
\def\sfk{\mathsf{k}}

\newcommand{\blue}[1]{{\color[rgb]{.3,.5,1}#1}}

\renewcommand{\blue}[1]{#1}
\usepackage[toc,page,header]{appendix}
\usepackage{minitoc}
\usepackage{silence}
\usepackage{bibunits}

\usepackage{hyperref}
\title{Improved Regret Analysis for Variance-Adaptive Linear Bandits and Horizon-Free Linear Mixture MDPs}

\author{%
\And
  Yeoneung Kim\thanks{Work done while at Seoul National University.}\\
  Gachon University\\
  \texttt{yeoneung@gachon.ac.kr} \\
\And
  Insoon Yang\\
  Seoul National University\\
  \texttt{insoonyang@snu.ac.kr} \\
\And
  Kwang-Sung Jun\\
  University of Arizona\\
  \texttt{kjun@cs.arizona.edu} \\
}

\begin{document}
%\doparttoc % Tell to minitoc to generate a toc for the parts
%\faketableofcontents % Run a fake tableofcontents command for the partocs
%\part{} % Start the document part
%\parttoc % Insert the document TOC

%----------- for separate bib
%\begin{bibunit}[plainnat]

\setlength{\abovedisplayskip}{4pt}
\setlength{\belowdisplayskip}{3pt}
\setlength{\abovedisplayshortskip}{4pt}
\setlength{\belowdisplayshortskip}{3pt}

%-- for reducing space after algorithm environment
% \textfloatsep=.5em

\maketitle

\begin{abstract}
  In online learning problems, exploiting low variance plays an important role in obtaining tight performance guarantees yet is challenging because variances are often not known a priori.
  Recently, considerable progress has been made by Zhang et al. (2021) where they obtain a variance-adaptive regret bound for linear bandits without knowledge of the variances and a horizon-free regret bound for linear mixture Markov decision processes (MDPs).
  In this paper, we present novel analyses that improve their regret bounds significantly.
  For linear bandits, we achieve $\tilde O(\min\{d\sqrt{K}, d^{1.5}\sqrt{\sum_{k=1}^K \sigma_k^2}\} + d^2)$ where $d$ is the dimension of the features, $K$ is the time horizon, and $\sigma_k^2$ is the noise variance at time step $k$, and $\tilde O$ ignores polylogarithmic dependence, which is a factor of $d^3$ improvement.
  For linear mixture MDPs with the assumption of maximum cumulative reward in an episode being in $[0,1]$, we achieve a horizon-free regret bound of $\tilde O(d \sqrt{K} + d^2)$ where $d$ is the number of base models and $K$ is the number of episodes.
  This is a factor of $d^{3.5}$ improvement in the leading term and $d^7$ in the lower order term.
  Our analysis critically relies on a novel peeling-based regret analysis that leverages the elliptical potential `count' lemma. 
\end{abstract}

\section{Introduction}
\def\VAR{\text{VAR}}

In online learning, variance often plays an important role in achieving low regret bounds.
For example, for the prediction with expert advice problem, \citet{hazan10extracting} proposed an algorithm that achieves a regret bound of $O(\sqrt{\text{VAR}_K})$ where $\text{VAR}_K$ is a suitably-defined variance of the loss function up to time step $K$, without knowing $\text{VAR}_K$ ahead of time.
The implication is that when the given sequence of loss functions has a small variance, one can perform much better than the previously known regret bound $O(\sqrt{K})$. 
For multi-armed bandits, \citet{audibert2006use} proposed an algorithm that achieves regret bounds that depends on the variances of the arms, which means that, again, the regret bound becomes smaller as the variances become smaller.

It is thus natural to obtain similar variance-adaptive bounds for other problems.
For example, in $d$-dimensional stochastic contextual bandit problems, the optimal worst-case regret bound is $\tilde O(\sig d\sqrt{K})$ where $\tilde O$ hides polylogarithmic dependencies and $\sigma^2$ is a uniform upper bound on the noise variance.
Following the developments in other online learning problems, it is natural to ask if we can develop a similar variance-adaptive regret bound.
The recent work by~\citet{zhang21variance} has provided an affirmative answer.
Their algorithm called VOFUL achieves a regret bound of $\tilde O(d^{4.5}\sqrt{\sum_{k=1}^K \sig_k^2} + d^5)$ where $\sig_k^2$ is the (unknown) noise variance at time step $k$.
This implies that, indeed, it is possible to adapt to the variance and suffer a much lower regret.
Furthermore, they show that a similar variance-adaptive analysis can be used to solve linear mixture Markov decision processes (MDPs) with the \textit{unit cumulative rewards} assumption :
\begin{align}\label{eq:ucr}
  \sum_{h} r^k_h \in [0,1], \forall k  
\end{align}
where $r^k_h$ is the reward received at episode $k$ and horizon $h$.
They show a regret bound of $\tilde O(d^{4.5}\sqrt{K}+d^9)$, which does not depend on the planning horizon length $H$ up to polylogarithmic factors.
We elaborate more on the linear bandit and linear mixture MDP problems in Section~\ref{sec:prelim}.

However, the regret rates of these problems have a large gap between the known lower and the upper bounds.
For example, in linear bandits, it is well-known that the regret bound has to be $\Omega(d\sqrt{K})$~\cite{dani08stochastic}, which rejects the possibility of obtaining $o(d\sqrt{\sum_{k=1}^K \sig^2_k})$, yet the best upper bound obtained so far is $O(d^{4.5}\sqrt{\sum_{k=1}^K \sig^2_k})$.
Thus, the gap is a factor of $d^{3.5}$, which is quite large.

In this paper, we reduce such gaps significantly by obtaining much tighter regret upper bounds.
Specifically, we show that a slight variation of VOFUL~\cite{zhang21variance} for linear bandits has a regret bound of $\tilde O(\min\{d\sqrt{K}, d^{1.5}\sqrt{\sum_{k=1}^K \sig_k^2})\}$ without knowledge of the variances.
This reduces the gap between the upper and lower bounds to only $\sqrt{d}$ for the leading term in the regret.
Furthermore, we employ a similar technique to show that the algorithm VARLin~\cite{zhang21variance} for linear mixture MDPs with unit cumulative rewards has a regret bound of $\tilde O(d\sqrt{K} + d^2)$.
At the heart of our analysis is a direct peeling of the instantaneous regret terms using an elliptical potential `count' lemma (EPC).
EPC bounds, given $q>0$, how many times $\|x_k\|_{V_{k-1}^{-1}}^2 \geq q$ happens from time $k=1$ to $\infty$ where $V_{k-1} = \sum_{s=1}^{k-1} x_s x_s^\T$.
Our lemma is an improved and generalized version of~\cite[Exercise 19.3]{lattimore20bandit}, which was originally used for improving the regret bound of linear bandit algorithms.
%, but a similar lemma appears earlier in \cite[Proposition 3]{russo13eluder}.
%We believe both our peeling-based analysis and the elliptical potential count lemma can be of independent interest.
We provide the proofs of our main results for linear bandits and linear mixture MDPs in Section~\ref{sec:lb} and Section~\ref{sec:lmmdp} respectively.
Finally, we conclude the paper with exciting future directions.

\textbf{Related work.~}
There are numerous works on linear bandit problems such as \cite{dani08stochastic, auer03nonstochastic,ay11improved,li19nearly} where the information of variance is not used. On the other hand, variance can be exploited to obtain better regret \cite{audibert2006use}. Recently, works by \cite{zhang21variance, zhou2021nearly} proposed ways to infuse the variance information in the regret analysis which improves the standard regret bound. 
Reinforcement learning with linear function approximation has been widely studied to develop efficient learning methods that work for large state-action space \cite{yang2019sample,wen2013efficient,jiang2017contextual,du2019provably,jin2020provably,wang2020reward,wang2020provably,wang20optimism,zanette2020learning,misra2020kinematic,krishnamurthy2016pac,dann2018oracle,sun2019model,feng2020provably,du2020agnostic,yang2020reinforcement}. 
To our knowledge, all aforementioned works derived a regret bound that depends on the planning horizon $H$ polynomially. 
It was \citet{zhang21variance} who first remove the polynomial dependence of $H$ in the linear mixture MDP problem, achieving a bound of $\tilde O(d^{4.5}\sqrt{K}+d^9)$.
In contrast, our analysis shows that their algorithm in fact achieves significantly better bound of $\tilde O(d\sqrt{K} + d^2)$.
Note that these results assume the unit cumulative rewards assumption~\eqref{eq:ucr} and time-homogeneous transition models.
In a similar setup where $r_{h,k} \in [0,1]$ with time-inhomogeneous transition models, \citet{zhou2021nearly} achieve the regret $\tilde O(\sqrt{d^2H+dH^3}\sqrt{HK}+d^2H^3+d^3H^2)$ and show a lower bound of $\tilde \Omega(dH^{3/2}\sqrt{K})$.
These problem setups are incompatible to the setup of~\cite{zhang21variance} and ours.

\section{Problem Definition}
\label{sec:prelim}

\textbf{Notations.~}
We denote $d$-dimensional $\ell_2$ ball by ${\BB_2^d(R)} := \{x \in \RR^d: \|x\|_2 \leq R\}$ and define $\BB_1^d(R)$ similarly for the $\ell_1$ ball.
Let ${[N]} := \{1,2,.\ldots,N\}$ for $N\in \mathbb{N}$. 
Given $\ell\in\RR$ and $x\in\RR$, we define the clipping operator as follows (take $0/0 = 0$):
\begin{align}\label{eq:clipping}
    {\clip{x}_\ell} := \min\cbr{|x| , 2^{-\ell} } \cd \fr{x}{ |x| } ~.
\end{align}

\textbf{Linear bandits.~}
The linear bandit problem has the following protocol.
At time step $k$, the learner observes an arm set $\blue{\mathcal{X}_k} \subseteq \mathbb{B}_2^d(1)$, chooses an arm $x_k \in \mathcal{X}_k$, pulls it.
The learner then receives a stochastic reward 
$r_k = x_k^\T \th^*+\eps_k$ where $\th^* \in \mathbb{B}_2^d(1)$ is an unknown parameter and $\eps_k$ is a zero-mean stochastic noise.
Following~\cite{zhang21variance}, we assume that $(i)$ $\forall k \in [K]$, $|r_k|\le [-\fr12,\fr12]$ almost surely, $(ii)$ $\mathbb{E}[\eps_k | \mathcal{F}_k]=0$ where $\mathcal{F}_k=\sigma(x_1,\eps_1,...,x_{k-1},\eps_{k-1}, x_k)$, and $(iii)$ $\mathbb{E}[\eps_k^2|\mathcal{F}_k]=\sigma_k^2$~.
Note that the bound on $|r_k|$ implies that $|\eps_k| \le 1$ almost surely.
Our goal is to minimize the regret %where %$\mathbb{E}[\mathcal{R}^K]$ where 
$   \mathcal{R}^K = \sum_{k=1}^K \max_{x\in \mathcal{X}_k} x^\T\th^*-x_k^\T \th^* ~.
$

\textbf{Linear mixture MDPs.~}
We consider an episodic Markov Decision Process (MDP) with a tuple $(\mathcal{S},\mathcal{A},r(s,a),P(s'|s,a),K,H)$ where $\mathcal{S}$ is the state space, $\mathcal{A}$ is the action space, $r:\mathcal{S}\times \mathcal{A} \rightarrow [0,1]$ is the reward function, $P(s'|s,a)$ is the transition probability, $K$ is the number of episodes, and $H$ is the planning horizon.
A policy is defined as $\pi = \{\pi_h:\mathcal{S}\rightarrow \mathcal{D}(\mathcal{A})\}_{h=1}^H$ where $\mathcal{D}(\mathcal{A})$ is a set of all distributions over $\mathcal{A}$.
For each episode $k\in[K]$, the learner chooses a policy $\pi^k$, and then the environment executes $\pi^k$ on the MDP by successively following $a_h^k \sim \pi_h^k(s_h^k)$ and $s_{h+1}^k \sim P(\cd|s_h^k,a_h^k)$.
Then, the learner observes the rewards $\{r_h^k \in [0,1]\}_{k,h}$ and moves onto the next episode.
The key modeling assumption of linear mixture MDPs is that the transition probability $P$ is a linear combination of a known set of models $\{P^i\}$, namely, $P=\sum_{i=1}^d \th_i^* P^i$
where $\th^* \in \mathbb{B}_1^d(1)$ is an unknown parameter.
We follow~\cite{zhang21variance} and make the following assumptions: 
\begin{itemize}
    \item The reward at each time step $h$ and episode $k$ is $r_h^k = r(s_h^k,a_h^k)$ for some known function $r:\mathcal{S}\times\mathcal{A} \rightarrow [0,1]$. 
    \item Unit cumulative rewards:$\sum_{h=1}^H r_h^k \in [0,1]$ for any policy $\pi^k$.
\end{itemize}
For a policy $\pi$, $V_h^\pi(s):=\max_{a\in \mathcal{A}} Q_h^\pi(s,a)$ where $Q_h^\pi(s,a)=r(s,a)+\mathbb{E}_{s'\sim P(\cd|s,a)}V_{h+1}^\pi(s')$ and $V_{H+1}^\pi(s) := 0$. Denoting $V^\pi(s_1)=V_1^\pi(s_1)$ and $V^*(s_1)=V^{\pi^*}(s_1)$, our goal is to minimize the regret 
\begin{equation*}
    \mathcal{R}^K =\sum_{k=1}^K V^*(s_1^k)-V^k(s_1^k) ~.
\end{equation*}

\section{Variance-Adaptive Linear Bandits}
\label{sec:lb}

In this section, we show that VOFUL of \citet{zhang21variance} has a tighter regret bound than what was reported in their work.
Our version of VOFUL, which we call VOFUL2, has a slightly different confidence set for ease of exposition.
Specifically, we use a confidence set that works for every $\mu \in \mathbb{B}^d_2(2)$ rather than over an $\eps$-net of $\mathbb{B}^d_2(2)$ (but we do use an $\eps$-net for the proof of the confidence set).

The full pseudocode can be found in Algorithm~\ref{alg:VOFUL2}.
VOFUL2 follows the standard optimism-based arm selection~\cite{auer02using,dani08stochastic,ay11improved}.
Let $\blue{\eps_s(\th)} := r_s - x_s^\top\th$ and $\blue{\eps^2_s(\th)} := (\eps_s(\th))^2$.
With $L$ and $\iota$ defined in Algorithm~\ref{alg:VOFUL2}, we define our confidence set after $k$ time steps as 
\begin{align}\label{eq:voful2conf-0}
    \blue{\Theta_{k}} := \cap_{\ell=1}^L \Theta^\ell_k
\end{align}
where
\begin{align}
  &\blue{\Theta^\ell_k} := \Bigg\{ \th \in \BB_2^d(1): \lt|\sum_{s=1}^{k} \clip{x_s^\T \mu }_\ell \eps_s(\th)\rt| \notag
  \leq  \sqrt{\sum_{s=1}^{k} \clip{x_s^\T\mu }_\ell^2 \eps^2_s(\th) \iota } +  2^{-\ell}\iota , \forall \mu\in {\BB_2^d(2)}  \Bigg\}
\end{align}
and the clipping operator $\clip{z}_\ell$ is defined in~\eqref{eq:clipping}.
The role of clipping is two-fold: (i) it allows us to factor out $\sum_{s=1} \eps_s^2(\th)$ by $\sum_s \clip{x_s^\T\mu }_\ell^2 \eps^2_s(\th) \le (2^{-\ell})^2 \sum_{s=1} \eps^2_s(\th)$ and (ii) the lower order term is reduced to the order of $2^{-\ell}$.
Both properties are critical in obtaining variance-adaptive regret bounds as discussed in~\cite{zhang21variance}.
The true parameter is contained in our confidence set with high probability as follows.
\begin{lemma}\label{lem:confset}
  (Confidence set)
  Let $L$, $\iota$, and $\dt$ be given as those in Algorithm~\ref{alg:VOFUL2}.
  Then, 
  \begin{align*}
    \mathbb{P}(\blue{\mathcal{E}_1}:=\{\forall k\in[K], \th^* \in  \Theta_k \}) \geq 1-\dt~.
  \end{align*}
\end{lemma}
In fact, in our algorithm, we use the confidence set of $\cap_{s=1}^{k-1} \Theta_s$ at time step $k$ for a technical reason.
VOFUL2 has the following regret bound.
\begin{theorem}\label{main-thm-bandit}
  VOFUL2 satisfies, with probability at least $1-2\dt$, 
  \begin{align*}
    \mathcal{R}^K =\tilde O\del{d^{1.5} \sqrt{\sum_{k=1}^K \sig_k^2\ln(1/\dt)} + d^2 \ln(1/\dt)}
  \end{align*}
  where $\tilde O$ hides poly-logarithmic dependence on $\{d,K,\sum_{k=1}^K \sig_k^{2},  \ln(1/\dt)\}$. 
\end{theorem}
Note that one can also show that VOFUL2 can be slightly modified to achieve the regret bound of $\tilde O\del[1]{  \min\cbr[1]{d\sqrt{K\ln(1/\dt)}, d^{1.5} \sqrt{\sum_{k=1}^K \sig_k^2\ln(1/\dt)}} + d^2 \ln(1/\dt)} $, thus being no worse than OFUL.
We postpone the proof of this to Section~\ref{sec:dsqrtK} to avoid clutter.

\textfloatsep=.5em

\begin{algorithm}[t]
  \begin{algorithmic}[1]
    \STATE \textbf{Initialize:} ${L} = 1 \vee \lt\lfl \log_2(K ) \rt\rfl$ where $\iota =128\ln((12K2^L)^{d+2}/\dt)$ and ${\dt} \le e^{-1}$.
    \FOR {$k=1,2,\ldots,K$}
    \STATE Observe a decision set ${\mathcal{X}_k} \subseteq \BB_2^d(1)$.
    \STATE Compute the optimistic arm as following: $x_k = \arg \max_{x \in \mathcal{X}_k} \max_{\th \in \cap_{s=1}^{k-1} \Theta_{s}} x^\top \theta$
    where $\Theta_{s}$ is defined in~\eqref{eq:voful2conf-0}.
    \label{line:a}
    \STATE Receive a reward $r_k$.
    \ENDFOR
  \end{algorithmic}
  \caption{VOFUL2}
  \label{alg:VOFUL2}
\end{algorithm}

\textbf{Properties of the confidence sets and implications on the regret.}
Before presenting the proof of Theorem~\ref{main-thm-bandit}, we provide some key properties of our confidence set (Lemma~\ref{lem:concreg-emp}) and the intuition behind our regret bound.
First, let us describe a few preliminaries.
Define
\begin{equation*}
    \blue{W_{\ell,k-1}(\mu)} := 2^{-\ell}  I + \sum_{s=1}^{k-1} \del{1\wedge \fr{2^{-\ell}}{|x_s^\T\mu|}} x_s x_s^\top.
\end{equation*}
Let ${\th_k}$ be the maximizer of the optimization problem at line~\ref{line:a} of Algorithm~\ref{alg:VOFUL2} and define ${\mu_k} = \th_k - \th^*$.
For brevity, we use a shorthand of
\begin{equation*}
    \blue{W_{\ell,k-1}} := W_{\ell,k-1}(\mu_k) = 2^{-\ell}  I + \sum_{s=1}^{k-1} \del{1\wedge \fr{2^{-\ell}}{|x_s^\T\mu_k|}} x_s x_s^\top ~.
\end{equation*}

Finally, we need to define the following event regarding the concentration of the empirical variance around the true variance:
\begin{align*}
\blue{\mathcal{E}_2} &:= \left\{\forall k\in[K], \sum_{s=1}^{k} \eps^2_s(\th^*) \right. 
                 \le \left. \sum_{s=1}^{k} 8 \sig_s^2 +  4 \log(\fr{4 K(\log_2(K) + 2)}{\dt})\right\}~,
\end{align*}
which is true with high probability as follows.
\begin{lemma}\label{lem:confvar} We have $\PP(\mathcal{E}_2) \geq 1-\dt$
\end{lemma}
\begin{proof}
The proof is a direct consequence of Lemma~\ref{lem:du-lem10} in our appendix.
\end{proof}

Let $\blue{\ell_k}$ be the integer $\ell$ such that $x_k^\T\mu_k \in (2 \cd 2^{-\ell}, 2 \cd 2^{-\ell+1}]$ and define $\blue{A_{k}} := \sum_{s=1}^k \sig_s^{2}$.
Lemma~\ref{lem:concreg-emp} below states the properties of our confidence set. 
\def\const{\mathsf{const}}
\begin{lemma}
  \label{lem:concreg-emp}
  Suppose the events $\cE_1$ and $\cE_2$ are true.
  Then, for any $k$ with $\ell_k = \ell$,
  \begin{enumerate}[(i)]
    \item \label{item:concreg-emp-1} 
    For some absolute constant $c_1$,
    \begin{align*}
    \|\mu_k\|_{W_{\ell,k-1}}^2 
    &\le 2^{-\ell}\sqrt{128 A_{k-1} \iota}   + 11\cd 2^{-\ell}\iota \le c_1 2^{-\ell} (\sqrt{A_{k-1} \iota} + \iota ),
    \end{align*}

    \item \label{item:concreg-emp-2} There exists an absolute constant $c_2$ such that $x_k^\T \mu_k \le c_2  \|x_k\|^{2}_{W^{-1}_{\ell,k-1}} \del{ \sqrt{ A_{k-1} \iota} + \iota} $.
  \end{enumerate}
\end{lemma}
The key difference between Lemma~\ref{lem:concreg-emp} and the results of~\citet{zhang21variance} is that we use the norm notations, although the norm involves a rather complicated matrix $W_{\ell,k-1}$.
This opens up possibilities of analyzing the regret of VOFUL2 with existing tools such as applying Cauchy-Schwarz inequality and the elliptical potential lemma \cite{ay11improved, cesa2006prediction,lattimore20bandit}.
In particular, Lemma~\ref{lem:concreg-emp}\ref{item:concreg-emp-2} seems useful because if we had such a result with $W_{\ell,k-1}$ replaced by $V_{k-1}= \lam I + \sum_{s=1}^{k-1} x_s x_s^\T$, then we would have, ignoring the additive term $\iota$,
\begin{align*}
    x_k^\T \mu_k \le \|x_k\|^2_{{V}^{-1}_{k-1}} \sqrt{\sum_{s=1}^{k-1} \sigma_s^2 \iota} ~.
\end{align*}
Together with the optimism and the  standard elliptical potential lemma (see Section~\ref{subsec:proof-voful2} for details), this leads to 
\begin{align*}
    \cR^K 
    &\le \sum_{k=1}^K x_k^\T \mu_k
    \le c_2 \sum_{k=1}^K \|x_k\|^2_{{V}^{-1}_{k-1}} \sqrt{\sum_{s=1}^{k-1} \sigma_s^2 \iota}
    \leq c_2 \cd O(d\log(T/d))\cd \sqrt{\sum_{s=1}^{K} \sigma_s^2 \iota}~.
\end{align*}
Since $\iota$ is linear in $d$, we would get the regret bounded by the order of $d^{1.5}\sqrt{\sum_{k=1}^K \sig_k^2}$, roughly speaking.
However, the discrepancy between $W_{\ell,k-1}$ and $V_{k-1}$ is not trivial to resolve, especially due to the fact that Lemma~\ref{lem:concreg-emp}\ref{item:concreg-emp-2} has $\mu_k$ on both left and the right hand side.
That is, $\mu_k$ is the key quantity that we need to understand, but we are bounding $x_k \mu_k$ as a function of $\mu_k$.
The novelty of our analysis of regret is exactly at relating $W_{\ell,k-1}$ to $V_{k-1}$ via a novel peeling-based analysis, which we present below.

\subsection{Proof of Theorem \ref{main-thm-bandit}}
\label{subsec:proof-voful2}

Throughout the proof, we condition $\cE_1$ and $\cE_2$ where each one is true with probability at least $1-\dt$, as shown in Lemma~\ref{lem:confset} and ~\ref{lem:confvar} respectively.
For our regret analysis, it is critical to use Lemma~\ref{lem:epc} below, which we call the elliptical `count' lemma.
This lemma is a generalization of \citet[Exercise 19.3]{lattimore20bandit}, which was originally used therein to improve the dependence of the range of the expected rewards in the regret bound.
Similar lemmas have been used in parallel studies \cite{he2021logarithmic,wagenmaker2022first}.
In particular, \citet{he2021logarithmic} employ a lemma similar to elliptical potential count and peeling technique for the regret analysis for the linear MDP as well, which we compare in detail in Section~\ref{app:diff} due to space constraint.
We remark that a similar strategy appears in disguise in \citet[Proposition 3]{russo13eluder} as well.

%We remark that a similar result appears in \citet[Proposition 3]{russo13eluder} as well.
%More recently, \citet{he2021logarithmic} employ a lemma similar to elliptical potential count and peeling technique for the regret analysis for the linear MDP as well, which we compare in detail in Section~\ref{app:diff} due to space constraint.
% but there are significant differences in approaches. The details are addressed in Section~\ref{app:diff} of our appendix.  
\begin{lemma}
  \label{lem:epc}
  (Elliptical potential count)
  Let $x_1, \ldots, x_k\in\RR^d$ be a sequence of vectors with $\|x_s\|_2 \le X$ for all $s\in[k]$.
  Let ${V_k} = \tau I + \sum_{s=1}^{k} x_s x_s^\T$ for some ${\tau}>0$.
  Let ${J} \subseteq[k]$ be the set of indices where $\|x_s\|_{V_{s-1}^{-1}}^2 \ge q$.
  Then,
  \begin{align*}
    |J| \le \frac{2}{\ln(1+q)} d \ln \del{1 + \fr{2/e}{\ln(1+q)}\fr {X^2} {\tau}}~.
  \end{align*}
\end{lemma}
As the name explains, the lemma above bounds how many times $\|x_s\|_{V_{s-1}^{-1}}^2$ can go above a given value $q>0$, which is different from existing elliptical potential lemmas that bound the sum of $\|x_s\|^2_{V_{s-1}^{-1}}$.
Let ${\th_k}$ be the $\th$ that maximizes the optimization problem at line~\ref{line:a} of Algorithm~\ref{alg:VOFUL2}.
We start by the usual optimism-based bounds: due to $\cE_1$, we have %For the event in Lemma \ref{lem:confset}, 
\begin{align*}
    \mathcal{R}^K &=\sum_{k=1}^K (\max_{x\in \mathcal{X}_k}(x^\T \th^*-x_k^\T \th^*)) 
    \le \sum_{k=1}^K (\max_{x \in \mathcal{X}_k,\th \in \Theta_k} x^\T\th -x_k^\T\th^*)
    =   \sum_{k=1} x_k^\T(\th_k - \th^*) =\sum_{k=1} x_k^\T \mu_k~.
\end{align*}
We now take a peeling-based regret analysis that is quite different from existing analysis techniques: % for linear bandits.
\begin{align*}
    \cR^K &\le \sum_{k=1}^{K} x_k^\T (\th_k - \th^*)
    \le {2^{-L}K +  \sum_{\ell=1}^{L} 2^{-\ell+2} \sum_{k=1}^{K}
      \one\cbr{x_k^\T\mu_k \in \lparen 2\cd2^{-\ell},2\cd2^{-\ell+1} \rbrack}  }~,
\end{align*}
where $L$ is defined in Algorithm~\ref{alg:VOFUL2}.
%
%Let ${G_\ell} := \{s\in[K]: \ell_s = \ell\}$ and $G_\ell[k] := G_\ell \cap [k]$.
Given $\ell$ and $k$, let $\blue{n_{k,\ell}}$ be such that $\max_{v:k\le v\le K, \ell_v = \ell}|x_k^\T \mu_v|  \in \lparen 2^{-\ell+n},2^{-\ell+n+1}\rbrack$ if such $n$ satisfies $n\ge1$.
%{
%\color{red}
%I think it should be
%\begin{equation*}
%  \max_{v:k<v\le K}|x_k^\T \mu_v|  \in \lparen 2^{-\ell+n},2^{-\ell+n+1}\rbrack.
%\end{equation*}
%as given $\ell$, we might not have $v$ such that $\ell_v = \ell$ so there is no meaning taking maximum with respect to $v$
%}
%\kj{
%I think it should be
%\begin{equation*}
%  \max_{v:k\le v \le K, \ell_v = \ell}|x_k^\T \mu_v|  \in \lparen 2^{-\ell+n},2^{-\ell+n+1}\rbrack.
%\end{equation*}
%Note that Lemma 4.4(i) works only when we have $k$ such that $\ell_k = \ell$.
%}
Otherwise, set $n_{k,\ell}=0$, which means $\max_{v:k\le v\le K, \ell_v = \ell}|x_k^\T \mu_v|  \le 2^{-\ell+n+1}$ with $n=0$.
We then define $\blue{G_{\ell,n}} := \{s\in[K-1]:\ell_s=\ell,n_{s,\ell}=n\}$ and let $\blue{G_{\ell,n}[k]} := G_{\ell,n} \cap [k]$.
Then,
\begin{align*}
  &\sum_{k=1}^{K}\one\cbr{x_k^\T\mu_k \in \lparen 2\cd2^{-\ell},2\cd2^{-\ell+1} \rbrack}=\sum_{k=1}^{K}\one\cbr{\ell_k=\ell}
  \le 1 + \sum_n \sum_{s\in G_{\ell,n}} 1~.
\end{align*}
Letting $\blue{V_{\ell,n,k-1}}  := 2^{-\ell} I + \sum_{s\in G_{\ell,n}[k-1]} x_s x_s^\top$,
a comparison between two matrices $W$ and $V$ is given as follows for every $v\in \{k,\ldots,K\}$:
\begin{align}\label{eq:bandit-key}
  W_{\ell,k-1}(\mu_v)
  =2^{-\ell} I + \sum_{s=1}^{k-1} \del{1\wedge \fr{2^{-\ell}}{|x_s^\T\mu_v|}} x_s x_s^\top \notag
  &\succeq 2^{-\ell} I + \sum_{s\in G_{\ell,n}[k-1]} \del{1\wedge \fr{2^{-\ell}}{2^{-\ell+n+1}}} x_s x_s^\top \notag
  \\& \succeq c\cd 2^{-n} V_{\ell,n,k-1}~.
\end{align}
%To rewrite the quantities in terms of $V$, we take a close look at the peeling in the sum. 
For $k\in G_{\ell,n}[K-1]$ and $u=\argmax_{v:k\le v\le K, \ell_v = \ell} |x_k^\T \mu_v|$, we have 
\begin{align*}
  2^{-\ell+n}
  <|x_k\mu_u|
  &\le \|x_k\|_{W_{\ell,k-1}^{-1}(\mu_u)} \|\mu_u\|_{W_{\ell,k-1}(\mu_u)}
  \\&\le \|x_k\|_{W_{\ell,k-1}^{-1}(\mu_u)} \|\mu_u\|_{W_{\ell,u-1}(\mu_u)} \tag{$u \ge k$ }
  \\&\le c\sqrt{2^n} \|x_k\|_{V_{\ell,n,k-1}^{-1}} \sqrt{2^{-\ell}(\sqrt{A_K\iota}+\iota)}~.
       \tag{\eqref{eq:bandit-key} and Lemma~\ref{lem:concreg-emp}\ref{item:concreg-emp-1}}
\end{align*}
Consequently,
$
  \|x_k\|^2_{V_{\ell,n,k-1}^{-1}} \ge c\frac{2^{-\ell+n}}{\sqrt{A_K\iota}+\iota}
$.
Thus, using Lemma~\ref{lem:epc} with $\tau = 2^{-\ell}$,
\begin{align*}
    1+\sum_{n=0}^\ell \sum_{k\in G_{\ell,n} } 1 
  &\le 1+\sum_n \sum_{k\in G_{\ell,n} } \onec{\|x_k\|^2_{V_{\ell,n,k-1}^{-1}} \ge c\frac{2^{-\ell+n}}{\sqrt{A_K\iota}+\iota}}
  \\&\le 1+ c \sum_n 2^{\ell-n} (\sqrt{A_K\iota} +\iota)d\ln\del{1 + c 4^\ell(\sqrt{A_K \iota}+\iota)}    
  \\&\le c 2^{\ell} (\sqrt{A_K\iota} +\iota)d\ln\del{1 + c 4^\ell(\sqrt{A_K \iota}+\iota)}~.
\end{align*}
where we use the fact that $1/\ln(1+q) \le c/q$ for an absolute constant $c$ if $q$ is bounded by an absolute constant. 
Finally,
\begin{align*}
  \sum_{k=1}^K x_k^\T \mu_k   
&\le 2^{-L} K + c \sum_{\ell=1}^L 2^{-\ell} 2^{\ell} (\sqrt{A_K\iota} +\iota)d\ln\del{1 + c 4^\ell(\sqrt{A_K \iota}+\iota)}
\\&= 2^{-L} K + c L(\sqrt{A_K\iota} +\iota)d\ln\del{1 + c 4^L(\sqrt{A_K \iota}+\iota)}~.
\end{align*}
We choose ${L} = 1 \vee \lt\lfl \log_2(K) \rt\rfl$, which leads to 
$
\mathcal{R}^K  
\le 
  c \del{\sqrt{A_{K-1}\iota} + \iota} 
  d \ln^2\del{1 + cK^2(\sqrt{A_{K-1}\iota} + \iota)}
$.
This concludes the proof.

\section{Linear Mixture MDP}
\label{sec:lmmdp}

As linear bandits and linear mixture MDPs have quite a similar nature, we bring the techniques in our analysis of VOFUL2 to improve the regret bound of VARLin of \citet{zhang21variance}. 
A key feature of linear mixture MDP setting is that one can estimate the upper bound of the variance as it is a quadratic function of $\theta^*$ while linear bandits do not have a structural assumption on the variance.
Thanks to such a structural property, we obtain a slightly better dependence on the dimension $d$. 
The confidence set derived for our proposed algorithm is slightly different from that of VARLin as ours is defined with $\forall \mu \in \mathbb{B}_1^d(2)$ rather than an $\eps$-net.
Our version of VARLin, which we call VARLin2, is described in Algorithm~\ref{alg:VARLin2}. 
Given $s_h^k$ and $a_h^k$, let us define $P_{s_h^k,a_h^k}(V_{h+1}^k):=\mathbb{E}_{s'\sim P_{s_h^k,a_h^k}}[V_{h+1}^k(s')]$ and $\blue{x_{k,h}^m} := [P^1_{s_h^k,a_h^k}(V_{h+1}^k)^{2^m},...,P_{s_h^k,a_h^k}^d(V_{h+1}^k)^{2^m}]^\T$ and let $L$, $\iota$, and $\dt$ be given as define Algorithm~\ref{alg:VARLin2}. 
Let $\blue{\eps_{v,u}^m(\th)} := \th^\T x_{v,u}^m - (V_{u+1}^v(s_{u+1}^v))^{2^m}$ for $(v,u)\in [K]\times [H]$, $m \in \{0,1,...,L\}$ where $L$ is defined in Algorithm~\ref{alg:VARLin2}.

We construct our confidence set as 
\begin{align}\label{eq:varlin2conf-0}
  \Th_k := \bigcap_{m=0}^{L} \bigcap_{i\in[L]} \bigcap_{\ell\in[L]} \Th_k^{m,i,\ell}
\end{align}
where we define $\Th_k^{m,i,\ell}$ below, based on the data collected up to episode $k-1$ .
First, let
\begin{align*}
    \blue{\eta_{k,h}^m} &:=\max_{\th \in \Theta_{k}} \{\th^\T x_{k,h}^{m+1} - (\th^\T x_{k,h}^m)^2\} 
\\ \text{~~ and ~~} \blue{\cT_{k,h}^{m,i}} &:= \{(v,u)\in ([k]\times[H]) \cup (\{k\}\times[h]): 
   \eta_{v,u}^m \in ( 2^{-i},2^{1-i}])\}~.
\end{align*}
We naturally define 
$\blue{\mathcal{T}_{k,h}^{m,L+1}}:=\{(v,u)\in \mathcal{T}_{k,h}^{m,i}: \eta_{v,u}^m \le 2^{-L}\}$.
With $\iota$ defined in Algorithm~\ref{alg:VARLin2}, define 
\begin{align} %\label{confidence-rl}
  \blue{\Theta^{m,i,\ell}_{k-1}} %  \notag
  &:= \Bigg\{\th \in \BB_1^d(1): \lt|\sum_{(v,u)\in \mathcal{T}_{k-1,H}^{m,i}} \clip{(x_{v,u}^m)^\T \mu }_\ell \eps_{v,u}^{m}(\th)\rt|\le \notag
\\  &\hspace{80pt} 4\!\sqrt{\!\sum_{(v,u)\in \mathcal{T}_{k-1,H}^{m,i}}\! \clip{(x_{v,u}^m)^\T\mu }_\ell^2 \eta_{v,u}^m \iota } +  4 \cdot 2^{-\ell}\iota , \forall \mu\in \BB_1^d(2)  \Bigg\} 
\end{align}
%$$\blue{\mathcal{T}_{k,h}^{m,i}}:=\mathcal{T}_{k-1,H}^{m,i}\bigcup \{(v,u)\in \mathcal{T}_{k}^{m,i}:u\le h,\eta_{v,u}^m \in  (2^{-i},2^{1-i}]\}.$$
We show that the confidence set is correct w.h.p. in the following lemma.
\begin{lemma}\label{lem:confset-rl}
 (Confidence set for MDP)~~
  $
    \PP(\forall k\in[K], \th^* \in  \Theta_k ) \ge 1-\dt
  $.
\end{lemma}
The consequence is that the $Q$ values computed in VarLin2 is optimistic with high probability due to the following property:
\begin{lemma} \label{lem:optimism}
For every $k \ge1$, 
$
\th^*\in \Theta_k \implies \forall h,s,a:Q_h^k(s,a)\ge Q^*(s,a)
    % \mathbb{P}(\forall k,h,s,a:Q_h^k(s,a)\ge Q^*(s,a))\ge \mathbb{P}(\forall k:\th^* \in \Theta_k)
$.
\end{lemma}

\begin{algorithm}[t]
  \caption{VARLin2}
  \label{alg:VARLin2}
  \begin{algorithmic}[1]
    \STATE \textbf{Initialize:} $\blue{L}=\lt\lfl \log_2 HK\rt\rfl+1$, $\blue{\iota} = 3\ln((2HK)^{2(d+3)}/\dt)$, $\blue{\dt} \le e^{-1}$.
    \FOR {$k=1,2,\ldots,K$}
    \FOR {$h=H,...,1$}
    \STATE For each $(s,a)\in \mathcal{S}\times\mathcal{A}$, define ${Q_h^k(s,a)}=\min\{1,r(s,a)+\max_{\th \in \Theta_{k-1}} \sum_{i=1}^d \th_i P_{s,a}^i V_{h+1}^k\}$ where $\Theta_{k-1}$ is defined in Lemma~\ref{lem:confset-rl}
    \STATE For each state $s$, ${V_h^k(s)}=\max_{a\in\mathcal{A}}Q_h^k(s,a)$.
    \ENDFOR
    \FOR {$h=1,...,H$}
    \STATE Choose $a_h^k = \argmax_{a\in\mathcal{A}} Q_h^k(s_h^k,a)$. 
    \STATE Observe a reward $r_h^k$ and the next state $s_{h+1}^k$.
    \ENDFOR

    %\label{line:a}

    \ENDFOR
  \end{algorithmic}
\end{algorithm}
Now with the confidence set defined above we state our main result. 
\begin{theorem}\label{main-rl}
With probability at least 1-$\dt$,
\begin{align*}
    \mathcal{R}^K
      &=\sum_{k=1}^K [V^*(s_1^k)-V^k(s_1^k)]
    = \tilde O(d\sqrt{K\log^2(1/\dt)} + d^2\log (1/\dt))~.
\end{align*} 
where $\tilde O$ hides poly-logarithmic dependence on $\{d,K,H, \ln(1/\dt)\}$.
\end{theorem}
% \gray{
% We remark that this upper bound is not improvable as far as the worst-case regret is concerned.
% To see this, consider removing our unit cumulative reward assumption and instead assuming $r^k_h\in[0,1]$.
% In this setup, \cite{zhou2021nearly} show that the worst-case regret of any algorithm is $\Omega(dH\sqrt{T})$.
% One can imagine using VARLin2 under this setup by feeding the algorithm a scaled reward $r'_{k,h} = 1/H r^k_h$, so that the unit cumulative reward assumption holds with $r'_{k,h}$.
% Our regret bound w.r.t. $r'_{k,h}$ now implies the bound of $\sqrt{d\sqrt{K}}$...
% }

\subsection{Proof of Theorem~\ref{main-rl}}
The main idea of the proof is to infuse a peeling-based argument together with elliptical potential count lemma to both the planning horizon and episode. 
Noting that the regret of the predicted variance is controlled by the variance of variance, one can expect to reduce the total regret using this information, as done in~\cite{zhang21variance}. We begin by introducing relevant quantities that are parallel with those in linear bandits. 
Let us first introduce the following lemma.

\begin{lemma}\label{lem:epc2}
  Let $x_1, \ldots, x_T\in\RR^d$ be a sequence of vectors with $\|x_t\|_2 \le X$ for all $t\in[T]$.
  % where $q>0$.
  Let ${V_t} = \lam I + \sum_{s=1}^{t} x_s x_s^\T$.
  Let $0 = \tau_0 < \tau_1 < \tau_2 < \ldots < \tau_z = T$ where $\tau_i$ marks the last time step of the $i$-th block formed by $\{\tau_{i-1} + 1, \ldots, \tau_i\}$ for all $i\in\{0,\ldots,z\}$.
  Let $\anc(t)$ be the `anchor' of $t$, the last time step of the $i$-th block such that the $(i+1)$-th block contains $t$: $\anc(t) = \max\{\tau_i: i \in \{0,\ldots,z\},  \tau_i < t\}$.
  Let $r>0$.
  Define ${J} \subseteq[T]$ to be the set of indices $t$ such that $\norm{x_t}^2_{V_{\anc(t)}^{-1}} > r \norm{x_t}^{2}_{V_{t-1}^{-1}} $
  is true for the first time in the block containing $t$.
  Then,
  \begin{align*}
    |J| \le \frac{2}{\ln(r)} d \ln \del{1 + \fr{2/e}{\ln(r)}\frac{X^2} \lam}~.
  \end{align*}
\end{lemma}
%\gray{
%To keep track of episode-index pairs in a simple but intuitive way we define a $K$ by $H$ matrix $\tilde D=(a_{i,j})$ where $a_{i,j}=(i,j)$ denotes $j-th$ time step in $i-th$ episode. Rewriting $\tilde D^\top=[r_1^\top,r_2^\top,...,r_K^\top]$ where $r_i$ denotes the $i$-th row of $\tilde D$, we will use a vection \kj{??} $D=[r_1,...,r_K] \in \mathbb{R}^{KH}$ for our index set. 
%For example, $D[2H+4]=(3,4)$ indicates the 4-th element in the third episode. 
%Clearly, $|D|=KH=:A$. 
%Regarding each row of $\tilde D$ as a block one can naturally define $\anc(a)\in [A]$ as well for $a\in [A]$.
%\kj{define $\anc(a)$ for our case precisely here}
%}

%\kj{note: the indexing $a$ coincides with the action. }
To keep track of episode-horizon index pairs concisely, we use a \textit{flat index} $t \in [T]$ where $\blue{T} := HK$.
Specifically, an episode $k$ and a horizon $h$ corresponds to the flat index $t = (k-1)H + h$.
Let $\blue{\sft(k,h)} :=  (k-1)H + h$.
Let $\blue{\sfk(t)}$ and $\blue{\sfh(t)}$ be the mapping from $t$ to its corresponding episode and horizon index respectively so that $k = \sfk(\sft(k,h))$ and $h = \sfh(\sft(k,h))$.
By taking $\tau_k$ in Lemma~\ref{lem:epc2} as $\sft(k,H)$, we have that $\blue{\anc(t)} := \sft(t-1,H)$.
%We define $\anc(t)$ be the flat index $t'$ such that $k(t')
We define $\blue{\cT_t^{m,i}} := \{\sft(k',h'): (k',h') \in \cT_{k,h}^{m,i}\}$ and $\mu^m_t := \mu^m_{\sfk(t), \sfh(t)}$.
Similarly, we define $x^m_t$, etc., by replacing the subscript $k,h$ with $t$. 
Hereafter, any appearance of subscript $k,h$ can be replaced with $t$ such that $t = \sft(k,h)$ without changing the meaning.

Given $m$, $k$ and $h$, we define $\blue{\ell_{k,h}^m}$ as the integer $\ell$ such that $ (x_{k,h}^m)^\T\mu_{k,h}^m \in \lparen 2\cd 2^{-\ell}, 2\cd 2^{-\ell+1}\rbrack$ where $\mu_{k,h}^m:=\theta_{k,h}^m -\theta^*$ and $\th_t^m=\argmax_{\th \in \Th_{k-1}} |\{\th^\T x_t^{m+1}-(\th^\T x_t^{m})^2\}|$. 
For simplicity, we abbreviate $\ell_{k,h}^m$ by $\ell$. %\kj{by $\ell$?}
Define $\blue{W_t^{m,i,\ell}(\mu)} 
    := 2^{-\ell}  I + \sum_{s\in \mathcal{T}_{t}^{m,i}} \del{1\wedge \fr{2^{-\ell}}{{|(x_s^{m})^\T\mu|}}} x_s^m (x_s^m)^\top$
and introduce a shorthand $\blue{ W_{\anc(t)}^{m,i,\ell}} := W_{\anc(t)}^{m,i,\ell}(\mu_{t}^m)~$ as before. With the definition above we have the following:
\begin{align*}
    2^{-\ell}  \|\mu_{t}^m \|^{2} + &\sum_{s\in \mathcal{T}_{\anc(t)}^{m,i}} \clip{(x_{s}^m)^\top\mu_{t}^m}_\ell (x_{s}^m)^\top\mu_{t}^m 
    =  \|\mu_{t}^m\|^2_{W_{{\anc}(t)}^{m,i,\ell}}~.
\end{align*}
We now show the key result of the confidence set of VarLin2 that parallels Lemma~\ref{lem:concreg-emp} for bandits.
\begin{lemma}\label{lem:concreg-emp-rl}
 Fix $m\in\{0,\ldots,L_\sfm\}$ and $i\in[L]$.
 Let $t \in \cT^{m,i}_T$.
 Then, with $\ell = \ell^m_t$,
% \label{item:concreg-emp-1-rl}
       \begin{align*}
        \|\mu_{t}^m\|_{W_{\anc(t)}^{m,i,\ell}}^2 &\le c_M \cd  \max\{\sqrt{2^{-i}\iota},\sqrt{2^{-\ell}\iota}\}
        \end{align*}
      where $c_M>0$ is an absolute constant.
    %\item \label{item:concreg-emp-1s-rl}
    %$\forall s\le k$ such that $\ell_s=\ell_k$, $\|\mu_{k,h}^m \|_{W_{i,\ell,s}}^2 
    %\le c \cd 2^{-\ell} (\sqrt{|\mathcal{T}_{k}^{m,i}|\cd 2^{-i}\iota}+\iota)$,
%    \item \label{item:concreg-emp-2-rl} There exists an absolute constant $c$ such that $(x_{k,h}^m)^\T \mu_{k,h}^m \le c\|{x_{k,h}^m}\|_{{W_{i,\ell,k}^h}^{-1}}^2 \cd (\sqrt{|\mathcal{T}_{k}^{m,i}|\cd 2^{-i}\iota}+\iota)$.
\end{lemma}
What is different from the linear bandit problem is that we do not update $\th$ until the planning horizon is over and an additional layer for peeling is imposed on variance. 
In \cite{zhang21variance}, the authors introduce an indicator $I_h^k$ to characterize episode-horizon pairs for which growth of the norm of $\mu$ with respect to $W_{t}$ is controlled by the norm with respect to $W_{\anc(t)}$ with $d^2$ growth rate, i.e., 
\begin{align*}
  I_h^k:= \one\{ \|\mu_t\|_{W_{t}^{m,i,\ell}} \le 4(d+2)^2 \|\mu_t\|_{W_{\anc(t)}^{m,i,\ell}}\}~.
\end{align*}
where $t=\sft(k,h)$.
Our novelty novelty lies in being able to replace $4(d+2)^2$ above by a constant rate $r$ that is set to 2 later (modulo some differences due to technical reasons).
To distinguish  we denote such a set by $I_{k,h}$. See \ref{proof_of_lem_I} for the definition of $I_{k,h}$ and the proof of the following lemma.
\begin{lemma}\label{lem-I}
  $\sum_{k=1}^K\sum_{h=1}^{H-1} I_{k,h} - I_{k,h+1}  \le  O(\fr{d}{\ln(r)}) \log(dHK(1+d^2/\ln(r)))$ for $r>0$.
\end{lemma}
Note here that once we fix $r$ such as $r=2$, the bound can be replaced by $O(d\log^5(dHK))$.
We now use the following regret decomposition due to~\cite{zhang21variance} which just come from replacing $I_h^k$ by $I_{k,h}$.
\begin{lemma}(\citet{zhang21variance})\label{lem:decomp}
$
    \mathcal{R}^K\le \Reg_1 +\Reg_2 + \Reg_3 + 
    \sum_{k=1}^K\sum_{h}^{H-1} (I_{k,h} - I_{k,h+1})
$
where 
$\Reg_1 =\sum_{k,h} (P_{s_h^k,a_h^k}V_{h+1}^k-V_{h+1}^k(s_{h+1}^k))I_{k,h}$, $\Reg_2 = \sum_{k,h}(V_h^k(s_h^k)-r_h^k-P_{s_h^k,a_h^k}V_{h+1}^k)I_{k,h}$, and $\Reg_3=\sum_{k=1}^K (\sum_{h=1}^H r_h^k-V_1^{\pi_k}(s_1^k))$.
\end{lemma}
Let $\blue{\Breve x_{k,h}^m} :=x_{k,h}^m I_{k,h}$ and define
$R_m$, $M_m$ as
\begin{align*}
    \blue{R_m} &:= \sum_{k,h} (\Breve x_{k,h}^m)^\T \mu_{k,h}^m,\quad\text{and}\quad
    \blue{M_m}:=\sum_{k,h} (P_{s_h^k,a_h^k}(V_{h+1}^k)^{2^m}-(V_{h+1}^k(s_{h+1}^k))^{2^m})I_{k,h}.
\end{align*} 
We have that $\Reg_1 = M_0$ and $\Reg_2 \le R_0$ since
\begin{align*}
    Q_h^k(s,a)-r(s,a)-P_{s,a}V_{h+1}^k\le \max_{\theta\in\Theta_k} x_{k,h}^0(\theta-\theta^*).
\end{align*}
To proceed, we first note that $\sum_{k,h} (I_{k,h} - I_{k,h+1})$ and $\Reg_3$ are bounded by  
$O(d \log^5(dHK))$ and $ O(\sqrt{K\log(1/\dt)})$ respectively from Lemma~\ref{lem-I} and Lemma~\ref{lem-R3}.
Since $\Reg_1 + \Reg_2 \le R_0 +M_0$, it remains to find a bound on $R_0+M_0$.
This, however, involves solving a series of recursive inequalities. % with multiple variables.
We leave the details in the appendix and provide a high-level description below.

Let us begin with Lemma~\ref{lem-M} in the appendix that shows
\begin{align}\label{eq:recursive}
 |M_m| \le \tilde O(\sqrt{M_{m+1}+ d +2^{m+1}(K\!+\!R_0)\log(1/\delta)}+\log(1/\delta))
\end{align}
where the RHS is a function of $\sqrt{M_{m+1}}$ and $\sqrt{R_0}$. 
Taking Proposition~\ref{prop:rl-bound} below for granted and combining it with the relation from \citet[Eq. (57)]{zhang21variance} showing 
\begin{align*}
    \sum_{k,h} \eta^m_{k,h} I_{k,h} &\le M_{m+1}+O(d\log^5(dHK))+2^{m+1}(K+R_0))+R_{m+1}+2R_m,    
\end{align*} 
one arrives at
$
    R_m\le \tilde O\big(d^{1/2}\!\!   \sqrt{(M_{m+1}\!+\!2^{m+1}(K\!+\!R_0)\!+\!R_{m+1}+2R_m}\!+\!d  )
$.
This bound is the key improvement we obtain via our peeling-based regret analysis.
Specifically, the bound on $R_m$ obtained by \cite{zhang21variance} has $d^4$ and $d^6$ in place of $d^{1/2}$ and $d$ above.

We first show how our regret bound helps in obtaining the stated regret bound and then present Proposition~\ref{prop:rl-bound}.
Noting that both $R_{L}$ and $M_{L}$ are trivially bounded by $HK$, one can solve the series of inequalities on $R_m$ and $|M_m|$ to obtain a bound on $R_0$:
\begin{align}%\label{eq:recursive}
    R_0 \le \tilde O\big(d^2\log (1/\dt)+\sqrt{d^2(K+R_0)\log(1/\dt)}\big).
\end{align}
Solving it for $R_0$, we obtain $
    R_0\le \tilde O\big(d^2\log (1/\dt)+\sqrt{d^2K\log(1/\dt)}\big)
$.
One can now plug in $R_0$ to the bound~\eqref{eq:recursive} and obtain a bound on $|M_0|$ in a similar way as follows, which concludes the proof:
$$
|M_0|\le \tilde O(d\sqrt{K\log^2(1/\dt)}+d^{2}\log(1/\dt)).
$$

We now show the key proposition that allows us to improve the bound on $R_m$. In the paper by \citet{zhang21variance}, $d^4$ was derived while we propose the following.
\begin{proposition}\label{prop:rl-bound}
Let $\blue{\breve\eta_{k,h}^m} := \eta_{k,h}^m I_{k,h}$.
Then, we have
\begin{align*}
R_m&\le O(d^{0.5}\log^{2.5}(HK)\sqrt{(1+\sum_{k,h}\Breve \eta_{k,h}^m) \iota \log(d\iota )}+ d \log^3(HK)\iota \log(d\iota))
\end{align*} 
\end{proposition}
\begin{proof}
Define $\blue{\mathcal{T}^{m,i,\ell}}:=\{t\in \mathcal{T}_{T}^{m,i} : (x_{t}^m)^\T \mu_{t}^m \in (2\cd2^{-\ell},2\cd 2^{1-\ell}]\}$
and split the time steps $\cT^{m,i,\ell}$ by
\begin{align*}
  \blue{\cT^{m,i,\ell,\la1\ra}} := \cbr{t\in \cT^{m,i,\ell}: \norm{\mu^m_{k,h}}_{W^{m,i,\ell}_{\anc(a)}} \le c_M \sqrt{2^{-i} \iota} }
  \text{~~ and ~~} \blue{\cT^{m,i,\ell,\la2\ra}} := \cT^{m,i,\ell}  \sm \cT^{m,i,\ell,\la1\ra}~.
\end{align*}
Having defined $I_{k,h}$ with $r=2$, we also denote $\blue{\brcT^{m,i,\ell,\la z\ra}} := \cT^{m,i,\ell,\la z\ra} \cap \cbr{t\in[T]: I_{t} = 1}$.
Now we decompose $R_m$ as
\begin{align*}
    R_m = \sum_{t\in[T]}  (\brx_{t}^m)^\T \mu_{t}^m
   = \sum_{i,\ell} \sum_{t \in \brcT^{m,i,\ell,\la1\ra}}  (\brx_{t}^m)^\T \mu_{t}^m
  + \sum_{i,\ell} \sum_{t \in \brcT^{m,i,\ell,\la2\ra}} (\brx_{t}^m)^\T \mu_{t}^m.
\end{align*}
Fix $m$, $i$ and $\ell$ and focus on $\sum_{t \in \brcT^{m,i,\ell,\la z\ra}} (\brx_{t}^m)^\T \mu_{t}^m $ for $z=1,2$.
Hereafter, we omit the superscripts and subscripts of $(m,i,\ell)$ to avoid clutter, unless there is a need. 
Note that for $t  \in \brcT^{\la1\ra,n}$ and $b$ such that $t<b \in \cT^{\la1\ra}$, 
\begin{align*}
  W_{\anc(t)}(\mu_b)
  = 2^{-\ell} I +  \sum_{t' \in \cT^{m,i,\ell}_{\anc(t)}} (1\wedge \fr{2^{-\ell} }{|x_{t'}^\T \mu_b | } )   x_{t'} x_{t'}^\T
  &\succeq 2^{-\ell} I +  \sum_{t' \in \cT^{\la1\ra,n}_{\anc(t)}  } (1\wedge \fr{2^{-\ell} }{2^{-\ell+n+1} } )   x_{t'} x_{t'}^\T
  \\&\succeq 2^{-\ell} I +  2^{-n-1} \sum_{t' \in \cT^{\la1\ra,n}_{\anc(t)} } x_{t'} x_{t'}^\T
  \\&\succeq c2^{-n} V^{\la1\ra,n}_{\anc(t)}~.
\end{align*}

For the same $t$, letting $b = \arg \max_{t\le b'\in \brcT^{\la1\ra} } |x_{t}^\T \mu_{b'}|$,
\begin{align*}
  2^{-\ell + n} 
  \le |x_{t}^\T \mu_{b}|
  \le \norm{x_{t}}_{W_{\anc(t)}^{-1}(\mu_{b})}  \norm{\mu_b}_{W_{\anc(t)}(\mu_b)}  
  &\le \sqrt{2^n} \norm{x_{t}}_{(V^{\la1\ra,n}_{\anc(t)})^{-1}}  \norm{\mu_{b}}_{W_{\anc(b)}(\mu_b)}  
%  \\&\le \norm{x_{k,h}}_{W_{v+1,h}^{-1}(\mu_{v,u})}  d^2 \norm{\mu_{v,u}}_{W_{u}(\mu_{v,u})}   \tag{by $(v,u) \in \brcT$}
  \\&\le c\sqrt{r} \sqrt{2^n} \norm{x_{t}}_{(V^{\la1\ra,n}_{t-1})^{-1}}  \sqrt{2^{-i}\iota}  \tag{by $b \in \brcT^{\la1\ra}$  }
\end{align*}
This implies that
$
  \norm{x_{t}}_{(V^{\la1\ra,n}_{t-1})^{-1}}^2  \ge c \fr{2^{-2\ell+n}}{r 2^{-i}\iota}
$.
Thus,
\begin{align*}
  \sum_{t\in \brcT^{\la1\ra}} x_{t}^\T\mu_{t}
  \le   c2^{-\ell} \sum_{t\in \brcT^{\la1\ra}} 1
  &\le c2^{-\ell} \sqrt{|\brcT^{\la1\ra}| \sum_{t\in \brcT^{\la1\ra}} 1}
  \le c2^{-\ell} \sqrt{|\brcT^{\la1\ra}| \sum_{n=0}^{\ell} \sum_{t\in \brcT^{\la1\ra,n}} 1}
  \\&\le c2^{-\ell} \sqrt{|\brcT^{\la1\ra}| \sum_{n=0}^{\ell} \sum_{t\in \brcT^{\la1\ra,n}} \onec{\norm{x_t }_{(V^{\la1\ra,n}_{t-1})^{-1}}^2 \ge c\fr{2^{-2\ell+n}}{r 2^{-i}\iota }} } 
  \\&\le c2^{-\ell} \sqrt{|\brcT^{\la1\ra}| \sum_{n=0}^{\ell} \sum_{t\in \cT^{\la1\ra,n}} \onec{\norm{x_t }_{(V^{\la1\ra,n}_{t-1})^{-1}}^2 \ge c\fr{2^{-2\ell+n}}{r 2^{-i}\iota }} } \tag{$\brcT^{\la1\ra,n} \subseteq \cT^{\la1\ra,n}$}
  \\&\le c2^{-\ell} \sqrt{|\brcT^{\la1\ra}| \sum_{n=0}^{\ell} \fr{r 2^{-i} \iota}{2^{-2\ell + n} } d \ln\del{ 1 + c\fr{r 2^{-i}\iota}{2^{-2\ell+n} 2^{-\ell} }  }  } 
  \\&\le c\sqrt{dr |\brcT^{\la1\ra}| 2^{-i} \iota \ln\del{ 1 + c r \iota 8^{\ell}  }  } 
    ~\le c \sqrt{dr (1 + \sum_{t \in \brcT^{\la1\ra}} \eta^m_{t} ) \iota \ln\del{ 1 + c r \iota 8^{\ell}  }  } 
\end{align*}
where we use the fact that $
    |\mathcal{T}^{\la1\ra}|\cd 2^{-i} \le O(1+\sum_{t \in \brcT^{\la1\ra}} \eta^m_{t})
$, which is straightforward by the definition. The summation over $\brcT^{(2)}$ can be handled in a similar way and the details of the proof is provided in Section \ref{proof_of_rl} in our appendix.
\end{proof}

\section{Conclusion}

In this work, we have made significant improvements in the regret upper bounds for linear bandits and linear mixture MDPs by employing a novel peeling-based regret analysis based on the elliptical potential count lemma. 
Our study opens up numerous future research directions.
First, the optimal regret rates are still not identified for these problems.  
It would be interesting to close the gap between the upper and lower bound.
Second, our algorithms are not computationally tractable.
We believe computationally tractable algorithms, even at the price of increased regret, may lead to practical algorithms.
Finally, characterizing variance-dependent uncertainty in the linear regression setting without prior knowledge of variances is an interesting statistical problem on its own.
Identifying novel estimators for it and proving their optimal coverage would be interesting.

% \section*{Acknowledgement}

\begin{ack}
The authors thank Liyu Chen for finding an error in our earlier version.
Insoon Yang is supported in part by the National Research Foundation of Korea (MSIT2020R1C1C1009766), the Information and Communications Technology Planning and Evaluation (IITP) grants (MSIT2022-0-00124, MSIT2022-0-00480), and Samsung Electronics.
Kwang-Sung Jun is supported by Data Science Academy and Research Innovation \& Impact
at University of Arizona.
\end{ack}

% is a fascinating 

% a very important and interesting problem in uncertainty quantification. 
% Identifying novel estimators for it and proving their optimal convergence properties are interesting statistical problems on their own.

%without unknown 
%performing linear regression while adapting to the unknown noise level is at the heart of our analysis. % is closely related to our confidence set.


%\putbib[ref]
%\end{bibunit}

\bibliographystyle{abbrvnat}
\bibliography{ref}

%%%%%%%%%%%%%%%%%%%%%%%%%%%%%%%%%%%%%%%%%%%%%%%%%%%%%%%%%%%%
\section*{Checklist}

\begin{enumerate}

  \item For all authors...
  \begin{enumerate}
    \item Do the main claims made in the abstract and introduction accurately reflect the paper's contributions and scope?
    \answerYes{}
    \item Did you describe the limitations of your work?
    \answerYes{}
    \item Did you discuss any potential negative societal impacts of your work?
    \answerNo{There is no negative impacts, to our knowledge.}
    \item Have you read the ethics review guidelines and ensured that your paper conforms to them?
    \answerYes{}
  \end{enumerate}

  \item If you are including theoretical results...
  \begin{enumerate}
    \item Did you state the full set of assumptions of all theoretical results?
    \answerYes{}
    \item Did you include complete proofs of all theoretical results?
    \answerYes{}
  \end{enumerate}

  \item If you ran experiments...
  \begin{enumerate}
    \item Did you include the code, data, and instructions needed to reproduce the main experimental results (either in the supplemental material or as a URL)?
    \answerNA{}
    \item Did you specify all the training details (e.g., data splits, hyperparameters, how they were chosen)?
    \answerNA{}
    \item Did you report error bars (e.g., with respect to the random seed after running experiments multiple times)?
    \answerNA{}
    \item Did you include the total amount of compute and the type of resources used (e.g., type of GPUs, internal cluster, or cloud provider)?
    \answerNA{}
  \end{enumerate}

  \item If you are using existing assets (e.g., code, data, models) or curating/releasing new assets...
  \begin{enumerate}
    \item If your work uses existing assets, did you cite the creators?
    \answerNA{}
    \item Did you mention the license of the assets?
    \answerNA{}
    \item Did you include any new assets either in the supplemental material or as a URL?
    \answerNA{}
    \item Did you discuss whether and how consent was obtained from people whose data you're using/curating?
    \answerNA{}
    \item Did you discuss whether the data you are using/curating contains personally identifiable information or offensive content?
    \answerNA{}
  \end{enumerate}

  \item If you used crowdsourcing or conducted research with human subjects...
  \begin{enumerate}
    \item Did you include the full text of instructions given to participants and screenshots, if applicable?
    \answerNA{}
    \item Did you describe any potential participant risks, with links to Institutional Review Board (IRB) approvals, if applicable?
    \answerNA{}
    \item Did you include the estimated hourly wage paid to participants and the total amount spent on participant compensation?
    \answerNA{}
  \end{enumerate}

\end{enumerate}

%%%%%%%%%%%%%%%%%%%%%%%%%%%%%%%%%%%%%%%%%%%%%%%%%%%%%%%%%%%%
%\bibliographystyle{abbrvnat}
%\bibliography{ref}
%
%\appendix
%
%
%\section{Appendix}
%
%
%Optionally include extra information (complete proofs, additional experiments and plots) in the appendix.
%This section will often be part of the supplemental material.

\appendix

%\addcontentsline{toc}{section}{Appendix} % Add the appendix text to the document TOC

\clearpage
\part{Appendix} % Start the appendix part

%\section{Appendix: example of $H_*^2$ and $\mathcal{C}_{\min}$} \label{Appendix: example of H and Cmin}
%\section{Appendix: Lower bound}\label{Appendix: lower bound}

\parttoc % Insert the appendix TOC
%\begin{bibunit}[plainnat]

\input{sub-appendix}
  
%  \putbib[ref]
%\end{bibunit}

%\bibliography{ref} % just so that my latex tool autocomplete the reference.

\end{document}

%% file: defs-kjun-v4a.tex
\usepackage{amsthm,amsmath,bbm,amsfonts,amssymb}
\usepackage{dsfont} % require for using \mathds{1}
\usepackage{mathtools}
\usepackage{commath}
\mathtoolsset{showonlyrefs=false}

\usepackage{xspace}
\usepackage[T1]{fontenc}
\usepackage[utf8]{inputenc}
\usepackage[usenames,dvipsnames]{xcolor} % for color names

\usepackage{hyperref,url}
\usepackage{cleveref} % must be loaded after amsthm

\usepackage[usenames,dvipsnames]{xcolor} % for color names

\usepackage{wrapfig}
\usepackage[export]{adjustbox}
\usepackage{algorithm}

\usepackage{graphicx,tabularx}
\usepackage{multirow,hhline}% http://ctan.org/pkg/multirow
\usepackage{booktabs}% to use \midrule for drawing a horizontal line in align environment

\usepackage{capt-of}

\allowdisplaybreaks % This allows breaking multi-line equations

\usepackage{pbox} % new line in a table cell -- http://tex.stackexchange.com/questions/2441/how-to-add-a-forced-line-break-inside-a-table-cell

\usepackage[export]{adjustbox}

\usepackage{pifont}% http://ctan.org/pkg/pifont
\usepackage{framed}
\usepackage[most]{tcolorbox}
\definecolor{kjgray}{rgb}{.7,.7,.7}

\newtheoremstyle{kjstyle}
{1ex} % Space above
{\topsep} % Space below
{\itshape} % Body font
{} % Indent amount
{\bfseries} % Theorem head font
{.} % Punctuation after theorem head
{.5em} % Space after theorem head
{} % Theorem head spec (can be left empty, meaning `normal')

\newtheoremstyle{kjstylenoitalic}
{1ex} % Space above
{\topsep} % Space below
{} % Body font
{} % Indent amount
{\bfseries} % Theorem head font
{.} % Punctuation after theorem head
{.5em} % Space after theorem head
{} % Theorem head spec (can be left empty, meaning `normal')

\tcbset{kjboxstyle/.style={title={},breakable,colback=white,enhanced jigsaw,boxrule=1.3pt,sharp corners,colframe=kjgray,boxsep=0pt,coltitle={black},attach title to upper={},left=.8ex,bottom=.4em}}    
\usepackage{mdframed}
\usepackage{lipsum}
\definecolor{kjgray}{rgb}{.7,.7,.7}
\makeatletter



\makeatletter
\renewcommand{\paragraph}{%
  \@startsection{paragraph}{4}%
  {\z@}{0.8em \@plus 1ex \@minus .2ex}{-1em}%
  {\normalfont\normalsize\bfseries}%
}
\makeatother

\newcounter{textcnt}

\newcommand\addtext[1]{%
  \stepcounter{textcnt}%
  \csgdef{text\thetextcnt}{#1}}

\newcounter{colnum}

\newcounter{Jidx}
\newcommand\dsymhelper[2]{
  \addtext{\hyperlink{#1}{#2}}%
  \blue{\hypertarget{#1}{#2}}%
}
\newcommand\dsym[1]{
  \stepcounter{Jidx}
  \xdef\tmpname{Jsym.\theJidx}
  \expandafter\dsymhelper\expandafter{\tmpname}{#1}
} 
\def\ddefloop#1{\ifx\ddefloop#1\else\ddef{#1}\expandafter\ddefloop\fi}
\def\ddef#1{\expandafter\def\csname c#1\endcsname{\ensuremath{\mathcal{#1}}}}
\ddefloop ABCDEFGHIJKLMNOPQRSTUVWXYZ\ddefloop
\def\ddef#1{\expandafter\def\csname b#1\endcsname{\ensuremath{{\boldsymbol{#1}}}}}
\ddefloop ABCDEFGHIJKLMNOPQRSUVWXYZabcdeghijklmnopqrtsuvwxyz\ddefloop  % including f here causes icml21 style file to behave weird.
\def\ddef#1{\expandafter\def\csname h#1\endcsname{\ensuremath{\widehat{#1}}}}
\ddefloop ABCDEFGHIJKLMNOPQRSTUVWXYZabcdefghijklmnopqrsuvwxyz\ddefloop % except for 't' because it is used by others
\def\ddef#1{\expandafter\def\csname hc#1\endcsname{\ensuremath{\widehat{\mathcal{#1}}}}}
\ddefloop ABCDEFGHIJKLMNOPQRSTUVWXYZ\ddefloop
\def\ddef#1{\expandafter\def\csname t#1\endcsname{\ensuremath{\widetilde{#1}}}}
\ddefloop ABCDEFGHIJKLMNOPQRSTUVWXYZabcdefghijklmnopqrtsuvwxyz\ddefloop
\def\ddef#1{\expandafter\def\csname r#1\endcsname{\ensuremath{\mathring{#1}}}}
\ddefloop ABCDEFGHIJKLMNOPQRSTUVWXYZabcdefghijklmnopqrtsuvwxyz\ddefloop
\def\ddef#1{\expandafter\def\csname tc#1\endcsname{\ensuremath{\widetilde{\mathcal{#1}}}}}
\ddefloop ABCDEFGHIJKLMNOPQRSTUVWXYZ\ddefloop

\DeclareMathOperator*{\argmax}{arg~max}

\DeclareMathOperator{\EE}{\mathbb{E}}
\DeclareMathOperator{\VV}{\mathbb{V}}
\DeclareMathOperator{\PP}{\mathbb{P}}

\DeclareMathOperator{\one}{\mathds{1}\hspace{-0.2em}}
\DeclareMathOperator{\Reg}{{\text{Reg}}}

\def\RR{{\mathbb{R}}}

\newcommand{\sr}[2]{ {\mathop{}\stackrel{#1}{#2}}\mathop{}\, }

\newcommand{\fr}[2]{ { \frac{#1}{#2} }}

\def\lt{\left}
\def\rt{\right}

\def\eps{\ensuremath{\epsilon}\xspace}

\def\T{\ensuremath{\top}}  % transpose
\def\sig{\ensuremath{\sigma}\xspace}

\def\eps{\ensuremath{\epsilon}\xspace}

\def\dt{{\ensuremath{\delta}\xspace} }
\def\sm{{\ensuremath{\setminus}\xspace} }

\def\lfl{\lfloor} 
\def\rfl{\rfloor}

\makeatletter
\newcommand{\vast}{\bBigg@{3}}
\newcommand{\Vast}{\bBigg@{4}}
\makeatother

\def\la{{\langle}}
\def\ra{{\rangle}}
\def\det{\ensuremath{\mbox{det}}}
\def\lam{\ensuremath{\lambda}}

\newcommand{\wbar}[1]{ {\ensuremath{\overline{#1}}} }

\def\hth{{\hat{ \theta}}}

\def\lam{{\ensuremath{\lambda}\xspace} }

\def\Th{{{\Theta}}}

\def\cF{{\ensuremath{\mathcal{F}}}}

\def\cR{\ensuremath{\mathcal{R}}}

\def\tr{{\ensuremath{\normalfont{\text{tr}}}}}

\def\th{{\ensuremath{\theta}}}

\def\cd{\cdot}

%% file: sub-appendix.tex
\section{Proofs for VOFUL2}

\subsection{Proof of Lemma \ref{lem:epc}}

\begin{proof}
  Let $W_t = V_0 + \sum_{s\in J,s\le t} x_s x_s^\T$.
  Then,
  \begin{align*}
    \del{\fr{d\tau + X^2|J|}{d} }^d
    &\ge \del{\fr{\tr(W_t)}{d} }^d 
    \\&\ge |W_t|   \tag{AM-GM ineq.}
    \\&= |V_0| \prod_{s\in J} (1 + \|x_s\|^2_{W_{s-1}^{-1} } ) \tag{rank-1 update equality for det.}
    \\&\ge |V_0| \prod_{s\in J} (1 + \|x_s\|^2_{V_{s-1}^{-1} } ) \tag{$W_{s-1} \preceq V_{s-1}$}
    \\&\ge \tau^d 2^{|J|}
    \\  \implies 
    |J| &\le \fr{d}{\ln(2)}\ln\del{1+\fr{X^2|J|}{d\tau} } 
  \end{align*}
  
  Let us generalize it so that we compute the number of times $\|x_s\|_{V_{t-1}^{-1}}^2 \ge q$ is true rather than $\|x_s\|_{V_{t-1}^{-1}}^2 \ge 1$ in which case we have
  \begin{align} \label{eq:20210425-01} 
    |J| &\le \fr{d}{\ln(1+q)}\ln\del{1+\fr{X^2|J|}{d\tau} } =: {A} \ln (1 + {B} |J|)
  \end{align}
  We want to solve it for $|J|$.
  We observe the following:
  \begin{align}
    |J| 
    \le A \ln(1 + B |J|) 
    &= A\del{\ln\del{\fr{|J|}{2A} } + \ln\del{2A(\fr{1}{|J|} + B) } } 
    \\&\le \fr{|J|}{2} + A \ln\del{\fr{2A}{e} \del{\fr{1}{|J|} + B } } 
    \\  \implies |J| &\le 2A \ln\del{\fr{2A}{e} \del{\fr{1}{|J|} + B } }  =  \fr{2}{\ln(1+q)} d \ln\del{\fr{2d}{e\ln(1+q)}\del{\fr 1 {|J|}  + \frac{ X^2}{ d\tau}}  }  \label{eq:20210425-03}
  \end{align}

  %
  %
  %With the usual technique, one can show that $\ln(a + b X) \le c X + f(X)$ for some $c$ and a function $f$, which leads to
  %\begin{align}\label{eq:20210425-2}
  %  |J| 
  %  &\le   \fr{2}{\ln(2)} d \ln\del{\fr{2}{e\ln(2)}\del{\fr d {|J|}  + \fr 1 \tau}  } 
  %  %  \\&\le cd \vee  \fr{2}{\ln(2)} d \ln\del{\fr{2}{e \ln(2)}\del{\fr 1 {c}  + \fr 1 \tau}  } \tag{case study with $|J|\le cd$ for any $c>0$}
  %  %  \\&\le \fr{2}{\ln(2)} d \ln\del{e + \fr{2}{e \ln(2) \tau}   } \tag{choose $c = \tfr{2}{e^2\ln(2)}$ }
  %\end{align}
  We fix $c>0$ and consider two cases:
  \begin{itemize}
    \item Case 1: $|J| < c d$ \\
    In this case, from~\eqref{eq:20210425-01}, we have $|J| \le \fr{d}{\ln(1+q)} \ln\del{1 + \fr{cX^2}{\tau} }  $ 
    \item Case 2: $|J| \ge cd$\\
    In this case, from~\eqref{eq:20210425-03} we have $ |J| \le  \fr{2}{\ln(1+q)} d \ln\del{\fr{2}{e \ln(1+q)}\del{\fr 1 {c}  + \fr {X^2} {\tau}}  }$
  \end{itemize}
  We set $c = \fr{2}{e \ln(1+q)}$ to obtain
  $   |J| \le \frac{2}{\ln(1+q)} d \ln \del{1 + \fr{2/e}{\ln(1+q)}\fr {X^2} {\tau}}  $.
  We remark that one can make the constant in front of the log to be $\frac{d}{\ln(1+q)}$ by plugging this bound into the RHS of~\eqref{eq:20210425-01}. 
\end{proof}

\subsection{Proof of Lemma \ref{lem:confset}}
\begin{proof}
  Let $\blue{\eps_s} := \eps_s(\th^*) = r_s - x^\T\th^*$.
  It suffices to show that the following is true w.p. at least $1 - \dt$, 
  \begin{align*}
      \forall \ell\in[L], k\in[K], \mu\in \cB_2^d(2), ~ \lt|\sum_{s=1}^{k} \clip{x_s^\T \mu }_\ell \eps_s\rt| \le  \sqrt{\sum_{s=1}^{k} \clip{x_s^\T\mu }_\ell^2 \eps^2_s \iota } +  2^{-\ell}\iota ~.
  \end{align*}
  To show this, we define ${\hcB_\ell}$ to be a ${\xi_\ell}$-net over $\mathbb{B}_2^d(2)$.
  with cardinality at most $\del{\fr{12}{\xi_\ell} }^d$.
  Such a net exists due to Lemma 4.1~\cite{pollard1990empirical}.
  Let us assume the following event, which happens with probability at least $1-6K\log_2(K)\sum_{\ell=1}^L |\hcB_\ell|$ by Lemma ~\ref{lem:du-thm4}:
  \begin{align}\label{eq:lem-conf}
    %\forall \ell\in[L], k\in[K],\mu'\in \hcB_\ell 
    \lt|\sum_{s=1}^{k} \clip{x_s^\T\mu' }_\ell \eps_s\rt| 
    \le 8\sqrt{\sum_{s=1}^{k} \clip{x_s^\T \mu' }_\ell^2 \eps_s^2 \ln(1/\dt)} +  16\cd 2^{-\ell}\ln(1/\dt)~. \tag{$\cE$}
  \end{align} 
  
  Let us fix $\ell \in [L]$, $k\in[K]$, and $\mu\in \mathbb{B}_2^d(2)$.
  Choose $\mu'\in \hcB_\ell$ such that $\|\mu - \mu'\|_2 \le\xi_\ell$.
  Then,
  \begin{align*}
    |\sum_{s=1}^{k} \clip{x_s^\T\mu} \eps_s|
    &\le  |\sum_{s=1}^{k} (\clip{x_s^\T\mu} - \clip{x_s^\T\mu'}) \eps_s| + |\sum_{s=1}^{k} \clip{x_s^\T\mu'} \eps_s|
    \\&\le \sum_{s=1}^{k} |\clip{x_s^\T\mu} - \clip{x_s^\T\mu'}| + |\sum_{s=1}^{k} \clip{x_s^\T\mu'} \eps_s| \tag{$|\eps_s|\le 1$}
    \\&\sr{(a)}{\le} k\xi_\ell + |\sum_{s=1}^{k} \clip{x_s^\T \mu'} \eps_s|
    \\&\le           k\xi_\ell + 8\sqrt{\sum_{s=1}^k \clip{x_s^\T\mu'}^2 \eps_s^2\ln(1/\dt)} + 16\cd 2^{-\ell}\ln(1/\dt)  \tag{by~\eqref{eq:lem-conf}}
    \\&\le k\xi_\ell + 8\sqrt{2\sum_{s=1}^k\del{ \clip{x_s^\T\mu}^2+\xi_\ell^2}\eps_s^2\ln(1/\dt)} + 16\cd 2^{-\ell}\ln(1/\dt)
    \\&\le k\xi_\ell +  \xi_\ell \cd 8\sqrt{2k\ln(1/\dt)} + 8\sqrt{2\sum_{s=1}^k \clip{x_s^\T\mu}^2\eps_s^2\ln(1/\dt)} + 16\cd 2^{-\ell}\ln(1/\dt)
    \\&\le           2^{-\ell}+2^{-\ell}\cd 8\sqrt{2\ln(1/\dt)} + 8\sqrt{2\sum_{s=1}^k \clip{x_s^\T\mu}^2\eps_s^2\ln(1/\dt)} + 16\cd 2^{-\ell}\ln(1/\dt)  \tag{choose $\xi_\ell = 2^{-\ell}/K$}
    \\&\le                       8\sqrt{2\sum_{s=1}^k \clip{x_s^\T\mu}^2\eps_s^2\ln(1/\dt)} + 32\cd 2^{-\ell}\ln(1/\dt) 
    \tag{by $1 \le \ln(1/\dt)$ }
  \end{align*}
  where $(a)$ follows from the fact that $|x_s^\top(\mu - \mu')|\le\eps$ and the observation that the clipping operation applied to two real values $z$ and $z'$ only makes them closer.
  It remains to adjust the confidence level.
  Note that
  \begin{equation*}
      \sum_{\ell=1}^L |\hcB_\ell|K = \sum_{\ell=1}^L (12K2^\ell)^dK\le 2(12K)^d \cd \frac{2^{Ld}}{2^d}\cd K \le (12K2^L)^{d+1}.
  \end{equation*}
  Thus,
  \begin{align*}
      6\log_2(K)\sum_{\ell=1}^L |\hcB_\ell|K \le (12K2^L)^{d+2}~.
  \end{align*}
  Replacing $\dt$ with $\dt/(12K2^L)^{d+2} $ and setting $\iota=128\ln((12K2^L)^{d+2}/\dt)$, we conclude the proof.
  We remark that we did not optimize the constants in this proof. 
\end{proof}

\subsection{Proof of Lemma \ref{lem:concreg-emp}}
\def\const{\mathsf{const}}
\begin{proof}
  Throughout the proof, every clipping operator $\clip{.}$ is a shorthand of $\clip{.}_\ell$. 
  For \ref{item:concreg-emp-1}, we note that
    \begin{align*}
    2^{-\ell}  \|\mu_k\|^{2} + \sum_{s=1}^{k} \clip{x_s^\top\mu_k}_\ell x_s^\top\mu_k 
    &= 2^{-\ell}  \|\mu_k\|^{2} + \sum_{s=1}^{k} \del{\del{1\wedge \fr{2^{-\ell}}{|x_s^\T\mu_k| } } x_s}^\top\mu_k x_s^\top\mu_k 
    \\&= \mu_k^\top\del{2^{-\ell}\lam I {+} \sum_{s=1}^{k} \del{1\wedge \fr{2^{-\ell}}{|x_s^\T\mu_k| } } x_s x_s^\top }\mu_k 
    \\&= \|\mu_k\|^2_{W_{\ell,k-1}}~.
  \end{align*} 
  Then,
  \begin{align*}
    &\|\mu_k\|_{(W_{\ell,k-1} - 2^{-\ell}  I )}^2 
    \\&= \sum_{s=1}^{k-1} \clip{x_s^\T\mu_k} x_s^\T\mu_k 
    \\&= \sum_{s=1}^{k-1} \clip{x_s^\T\mu_k} (x_s^\T\th_k - r_s + r_s - x_s^\T\th^*) 
    \\&= \sum_{s=1}^{k-1} \clip{x_s^\T\mu_k} (-\eps_s(\th_k) + \eps_s(\th^*)) 
    \\&\le  \sqrt{\sum_{s=1}^{k-1} \clip{x_s^\T\mu_k}^2 \eps^2_s(\th_k) \iota}  + 2^{-\ell} \iota + \sqrt{\sum_{s=1}^{k-1} \clip{x_s^\T\mu_k}^2 \eps_s^2(\th^*) \iota} + 2^{-\ell} \iota  
    \\&\sr{(a)}{\le} \sqrt{\sum_{s=1}^{k-1} \clip{x_s^\T\mu_k}^2 2(x_s^\T\mu_k)^2 \iota}  + 2 \sqrt{\sum_{s=1}^{k-1} \clip{x_s^\T\mu_k}^2 2\eps_s^2(\th^*) \iota} + 2 \cd 2^{-\ell} \iota
    \\&\le \sqrt{\sum_{s=1}^{k-1} \clip{x_s^\T\mu_k}^2 2(x_s^\T\mu_k)^2 \iota}  + 2^{-\ell}\sqrt{4\del{ \sum_{s=1}^{k-1} 8\sig_s^2 + 4 \ln(\fr{4 K (\log_2(K) + 2)}{\dt}) } \iota} + 2\cd2^{-\ell} \iota
        \tag{By $\cE_2$}
    \\&\le \sqrt{\sum_{s=1}^{k-1} \clip{x_s^\T\mu_k}^2 2(x_s^\T\mu_k)^2 \iota}  +  2^{-\ell}\sqrt{32 \sum_{s=1}^{k-1} \sig_s^2 \iota } + 3\cd 2^{-\ell} \iota 
    \\&\le \sqrt{4\sum_{s=1}^{k-1} 2^{-\ell} \clip{x_s^\T\mu_k} (x_s^\T\mu_k) \iota}  + 2^{-\ell}\sqrt{ 32 \sum_{s=1}^{k-1} \sig_s^2 \iota } + 3\cd 2^{-\ell} \iota          \tag{$|x_s^\T\mu_k |\le 2, \clip{\cd} \le 2^{-\ell}$  }
    \\&= \sqrt{4\cd 2^{-\ell} \|\mu\|_{(W_{\ell,k-1} - 2^{-\ell}  I )}^{2}  \iota}  + 2^{-\ell}\sqrt{32 \sum_{s=1}^{k-1} \sig_s^2 \iota } + 3\cd2^{-\ell} \iota
  \end{align*}
  where $(a)$ follows from $\eps^2_s(\th_k) = (r_s - x_s^\T\th_k)^2 = (x_s^\top(\th_* - \th_k) + \eps_s(\th^*))^2 \le 2(x_s^\T\mu_k)^2 + 2\eps_s^2$.
  We now have $\|\mu\|_{(W_{\ell,k-1} - 2^{-\ell}  I )}^{2}$ on both sides.
  Using $X \le A + \sqrt{B X} \le A + (B/2) + (X/2) \implies X \le 2A + B$, we have
  \begin{align*}
    \|\mu_k\|_{(W_{\ell,k-1} - 2^{-\ell}  I )}^2 
    &\le 2^{-\ell}\sqrt{128\sum_{s=1}^{k-1} \sig_s^2 \iota}   + 8\cd 2^{-\ell}\iota
    \\\implies     \|\mu_k\|_{W_{\ell,k-1}}^2 
    &\le 4\cd 2^{-\ell} + 2^{-\ell}\sqrt{128\sum_{s=1}^{k-1} \sig_s^2 \iota}   + {8}\cd 2^{-\ell}\iota.
  \end{align*}
  Since $1 \le \ln(1/\dt)$, we have $4 \cd 2^{-\ell}  \le 4 \cd 2^{-\ell} \ln(1/\dt) \le 2^{-\ell}\iota$, which concludes the proof of \ref{item:concreg-emp-1}.

  For \ref{item:concreg-emp-2}, let $c$ be an absolute constant that may be different every time it is used.
  We apply Cauchy-Schwarz inequality to obtain
  \begin{align*}
    (x_k^\T\mu_k)^2
    &\le \|x_k\|^2_{W_{\ell,k-1}^{-1}}  \|\mu_k\|^2_{W_{\ell,k-1}} 
    \\&\le \|x_k\|^2_{W_{\ell,k-1}^{-1}}  \cd c \cd \del{2^{-\ell}\sqrt{\sum_{s=1}^{k-1}\sig_s^2 \iota} + 2^{-\ell}\iota  }
    \\&\le \|x_k\|^2_{W_{\ell,k-1}^{-1}}  \cd c \cd x_k^\T\mu_k \del{\sqrt{\sum_{s=1}^{k-1}\sig_s^2 \iota} + \iota  }  \tag{$2^{-\ell}\le x_k^\T\mu_k \le 2^{-\ell + 1}$  }
  \end{align*}
  Dividing both sides by $x_k^\T\mu_k$ concludes the proof.
\end{proof}

\subsection{$d\sqrt{K}$ regret bound of VOFUL2}
\label{sec:dsqrtK}

Let us slightly modify the algorithm so we now add $\Theta_k^{\ell}$ with $\ell=0$:
\begin{align*}
  \Theta_k = \cap_{\ell=0}^{L} \Theta_k^\ell~.
\end{align*}
Let us call this algorithm VOFUL3.
Note that this slight change will not alter the order of the regret bound of VOFUL2 reported in Theorem~\ref{main-thm-bandit}.

Let $\lam>0$ be an analysis parameter to be determined later.
Let $X_k \in \RR^{k\times d}$ be the design matrix where row $s$ is $x_s^\T$ and define $y_k := (r_1,\ldots,r_k)^\T$, $\eta_k := (\eps_1,\ldots,\eps_k)^\T$.
Let $V_k := \lam I + X_k^\T X_k$ and 
\begin{align}\label{eq:ridge}
  \hth_k := V_k^{-1} X_k^\T y_k
\end{align}
We claim that
\begin{align*}
  \Theta_k  \subseteq \cbr{\th \in \BB_2^d(1):   \norm{\hth_k - \th}_{V_k}^2 \le \beta_k } =: \mathring{\Th}_k
\end{align*}
for some $\beta_k = \tilde O(d + \ln(1/\dt))$.
This suffices to show that the VOFUL2 has regret bound of $\tilde O(d\sqrt{K})$ since the proof technique of OFUL~\cite{ay11improved} can be immediately applied since the UCB computed based on $\Th_k$ is bounded above by the UCB computed with $\mathring{\Th}_k$.
Thus, VOFUL3 has regret bound of
\begin{align*}
  \mathcal{R}^K =\tilde O\del{d\sqrt{K \ln(1/\dt)},~ d^{1.5} \sqrt{\sum_{k=1}^K \sig_k^2\ln(1/\dt)} + d^2 \ln(1/\dt)} ~.
\end{align*}

To see why the claim above is true, let $\th \in \Theta_k$.
Then, using $\eps_s^2(\th) \le 4$, we have, $\forall \mu \in \BB_2^d(2)$,
\begin{align*}
  |\sum_s \mu^\T x_s(y_s - x_s^\T\th)| \le \norm{\mu}_{\sum_{s=1}^k x_s x_s^\T} \cd \sqrt{4\iota} + \iota
\end{align*}
Let us drop the subscript $k$ from $\cbr{\hth_k, V_k, X_k, y_k, \eta_k}$ for brevity.
The display above can be rewritten as
\begin{align*}
  \mu^\T (X^\T y - X^\T X \th)
  = \mu^\T V(\hth - \th) + \lam \mu^\T \th 
  &\le \norm{\mu}_{X^\T X} \cd \sqrt{4\iota}   + \iota
  \\\implies
  \mu^\T V(\hth - \th) \le \norm{\mu}_{V} \cd \sqrt{4\iota}   + \iota + 2\lam \tag{$\|\th\|_2 \le 1, |\mu^\T\th|\le 2$ }
\end{align*}
We can choose $\mu = \fr{1}{2}(\hth - \th)$ since $\fr{1}{2}\|\hth - \th\|_2 \le \fr12 (\|\hth\|_2 + \|\th\|_2)\le 2$ by Lemma~\ref{lem:rls_norm_bound}  and the choice of $\lam$ therein.
Then,
\begin{align*}
  \norm{\hth - \th}_V^2 &\le \norm{\hth - \th}_V \sqrt{4\iota}  + 2\iota + 4\lam
  \\ \implies
  \norm{\hth - \th}_V^2 &\le 8\iota + 8\lam  \tag{AM-GM ineq. on $\norm{\hth - \th}_V \sqrt{4\iota}$}
  \\&= \tilde O(d + \ln(1/\dt))
\end{align*}

\begin{lemma}\label{lem:rls_norm_bound}
  Take the assumptions for the linear bandit problem in Section~\ref{sec:prelim}. % except that $\|\th^*\|_2 \le S$ for some $S$.
  Consider $\hth_k$ defined in~\eqref{eq:ridge} with  $\lam = d\ln(1+\fr{K}{d}) + 2 \ln(1/\dt)$.
  Let $K \ge (e-1)d$.
  Then, with probability at least $1-\dt$, we have, 
  \begin{align*}
    \forall k \le K, \|\hth_k\|_2 \le 3.
  \end{align*}
\end{lemma}
\begin{proof}
  Let us drop the subscript $k$ from $\cbr{\hth_k, V_k, X_k, y_k, \eta_k}$.
  Then,
  \begin{align*}
    \|\hth\|_2
    &\le \|\hth - \th^*\|_2 + \|\th^*\|_2~.
  \end{align*}
  Note that $\norm{\th^*}_2 \le 1$.
  We can further bound the first term:
  \begin{align*}
    \|\hth - \th^*\|_2
    &=   \|V^{-1}(X^\T\eta - \lam \th^*) \|_2
    \\&\le \|V^{-1}X^\T \eta\|_2 + \lam \| V^{-1} \th^*\|_2~.
  \end{align*}
  Note that 
  \begin{align*}
    \lam \|V^{-1}\th^*\|_2 = \lam \sqrt{{\th^*}^\T V^{-2} \th^*} 
    \le \lam \sqrt{{\th^*}^\T (\fr1{\lam^2} I) \th^* } 
    \le \|\th^*\|_2 \le 1~.
  \end{align*}
  It remains to bound $ \|V^{-1}X^\T \eta\|_2$.
  To see this,
  \begin{align*}
    \|V^{-1}X^\T \eta\|_2 
    &= \sqrt{\eta^\T X V^{-2} X^\T \eta} 
    \\&= \sqrt{\eta^\T X V^{-1} (\fr1\lam \cd \lam I)  V^{-1} X^\T \eta} 
    \\&\le \sqrt{\eta^\T X V^{-1} (\fr1\lam \cd V)  V^{-1} X^\T \eta} 
    \\&= \fr{1}{\sqrt{\lam} } \|X^\T\eta\|_{V^{-1}}
    \\&\le \fr{1}{\sqrt{\lam} } \del{ \sqrt{d\ln(1+\fr{k}{d\lam}) + 2 \ln(1/\dt)} } \tag{\citet[Theorem 1]{ay11improved}}
    %    \\&\le \fr{1}{\sqrt{\lam} } \sqrt{d\ln(1+\fr{k}{d\lam}) + 2 \ln(1/\dt)}
    \\&\sr{(a)}{\le} 1
  \end{align*}
  where $(a)$ is by choosing $\lam = d\ln(1+\fr{K}{d}) + 2 \ln(1/\dt)$ and then using  $\lam \ge 1$ (due to $K \ge (e-1)d$).
\end{proof}

\subsection{Miscellaneous Lemmas}

For completeness, we state the lemmas borrowed from prior work.
\begin{lemma}\label{lem:du-thm4}(\citet[Lemma 9]{zhang21variance})
  Let $\left\{\mathcal{F}_{i}\right\}_{i=0}^{n}$ be a filtration. Let $\left\{X_{i}\right\}_{i=1}^{n}$ be a sequence of real-valued random variables such that $X_{i}$ is $\mathcal{F}_{i}$-measurable. We assume that $\mathbb{E}\left[X_{i} \mid \mathcal{F}_{i-1}\right]=0$ and that $\left|X_{i}\right| \le b$ almost surely. For $\delta<e^{-1}$, we have
  \begin{align*}
    \PP\del{|\sum_{i=1}^{^n} X_i| \le 8 \sqrt{\sum_{i=1}^{n}X_i^2 \ln(1/\dt)} + 16 b \ln(1/\dt) }  \ge 1 - 6\dt \log_2(n)
  \end{align*}
\end{lemma}

\begin{lemma}
  \label{lem:du-lem10}(\citet[Lemma 10]{zhang21variance})
  Let $\left\{\mathcal{F}_{i}\right\}_{i \ge 0}$ be a filtration. Let $\left\{X_{i}\right\}_{i=1}^{n}$ be a sequence of random variables such that $\left|X_{i}\right| \leq 1$ almost surely, that $X_{i}$ is $\mathcal{F}_{i}$-measurable. For every $\delta \in(0,1)$, we have
  $$
  \PP\left[\sum_{i=1}^{n} X_{i}^{2} \ge \sum_{i=1}^{n} 8 \mathbb{E}\left[X_{i}^{2} \mid \mathcal{F}_{i-1}\right]+4 \ln \frac{4}{\delta}\right] \leqslant\left(\left\lceil\log _{2} n\right\rceil+1\right) \delta
  $$
\end{lemma}

\section{Proofs for VARLin2}

Throughout the proof, we use $c$ as absolute constant that can be different every single time it is used.

\subsection{Proof of Lemma \ref{lem:optimism}}
\begin{proof}
Assume that $\th^* \in \Theta_k$ for all $k\in[K]$. Since $\th^* \in \Theta_k$,
\begin{align*}
    Q_h^k(s,a)&=\min\{r(s,a)+\max_{\th \in \Theta_k} \sum_{i=1}^d\th_iP_{s,a}^i V_{h+1}^k\}
    \\&\ge \min \{1,r(s,a)+\sum_{i=1}^d\theta_i^*P_{s,a}^iV_{h+1}^k\}
    \\&\ge \min \{1,r(s,a)+\sum_{i=1}^d\theta_i^*P_{s,a}^iV_{h+1}^*\}
    \\&=Q_h^*(s,a),
\end{align*}
so the statement follows.
\end{proof}

\subsection{Proof of Lemma \ref{lem:confset-rl} }

\begin{proof} 
\def\VV{\mathbb{V}}
\def\EE{\mathbb{E}}
\def\cF{\mathcal{F}}
Similar to the linear bandit case, let $\hcB_\ell$ be a $\xi_\ell$-net over $\mathbb{B}_1^d(2)$ with cardinality at most $(\frac{12}{\xi_\ell})^d$ and pick $\mu \in \mathbb{B}_1^d(2)$ and $\mu'\in \hcB_\ell$ such that the distance between them is at most $\xi_\ell$. Let us define the conditional variance $\VV_{v,u}[\eps^m_{v,u}] := \VV[\eps^m_{v,u} \mid \cF_{u}^v]$ where $\cF_{u}^v$ denotes history up to (and including) episode $v$ and time horizon $u$.
%Set $\eta_{k,h}^m = (\th^*)^\T x_{k,h}^{m+1}-((\th^*)^\T x_{k,h}^m)^2$ \kj{this notation conflicts with $\eta^m_{k,h}$ from the main text. Following E.3 of Zhang et al, we can just use $\VV[\eps^m_{v,u} \mid \cF_{u-1}^v]$ instead (todo: define $\cF_{u}^v$ as the $\sig$-algebra generated by the random variables up to (and including) episode $v$ and horizon $u$). Then, we can use $\VV_{v,u}[\eps^m_{v,u}] := \VV[\eps^m_{v,u} \mid \cF_{u}^v] \le ... \le \eta^m_{v,u}$ by Jensen's inequality.} and $\eps_{v,u}^m = (\theta^*)^\T x_{v,u}^m - (V_{u+1}^v (s_{u+1}^v))^{2^m}$.

% \kj{Q: did we define $\VV_{v,u}$?}
Noticing
\begin{align*}
  \VV_{v,u}[\eps^m_{v,u}] 
  &= \EE_{v,u}[(\eps^m_{v,u})^2]
\\&= ((\th^* )^\T x^{m}_{v,u})^2 - 2 ((\th^* )^\T x^{m}_{v,u}) \cd \EE_{v,u}[ (V^v_{u+1}(s^v_{u+1}))^{2^m} ] + \EE_{v,u}[ \del{(V_{u+1}^v(s^v_{u+1}))^{2^m} }^2],
\end{align*}
% Using Jensen's inequality, $((\th^*)^\T x^m_{v,u} )^2 \le (\th^*)^\T x^{m+1}_{v,u}$.
$\EE_{v,u}[ (V^v_{u+1}(s^v_{u+1}))^{2^m} ] = (\th^* )^\T x^{m}_{v,u}$, and $\EE_{v,u}[ \del{(V_{u+1}^v(s^v_{u+1}))^{2^m} }^2] = (\th^*)^\T x^{m+1}_{v,u}$, we have
$$\VV_{v,u}[\eps^m_{v,u}]  = (\th^*)^\T x^{m+1}_{v,u} - ((\th^*)^\T x^{m}_{v,u})^2.$$

%{
%\color{red}
%Zhang's paper
%$$\VV_{v,u-1}[\eps^m_{v,u}]  \le  (\th^*)^\T x^{m+1}_{v,u} - ((\th^*)^\T x^{m}_{v,u})^2 \le \eta_{v,u}^m$$
%}

We apply Lemma \ref{ineq} with $\eps=1$, $b=2^{-\ell}$ to obtain 
\begin{equation*}
    \lt|\sum_{(v,u)\in \mathcal{T}_{k,H}^{m,i}} \clip{(x_{v,u}^m)^\T \mu' }_\ell \eps_{v,u}^{m}\rt| \leq 4\sqrt{\sum_{(v,u)\in \mathcal{T}_{k,H}^{m,i}} \clip{(x_{v,u}^m)^\T\mu' }_\ell^2 \VV[\eps^m_{v,u} \mid \cF_{u}^v] } +  4 \cdot 2^{-\ell} \ln(1/\dt)
\end{equation*}
with probability at least $1-\delta(1+\log_2 (HK))$ and repeat the similar procedure by taking the union bound. 
We drop $\ell$ from the clipping notation for the sake of brevity. 
%  \begin{align*}
%    &|\sum_{(v,u)\in \mathcal{T}_{k,H}^{m,i}}  \clip{(x_{v,u}^m)^\T \mu } \eps_{v,u}^{m}|
%    \\&=   |\sum_{(v,u)\in \mathcal{T}_{k,H}^{m,i}} (\clip{(x_{v,u}^m)^\T \mu} - \clip{(x_{v,u}^m)^\T\mu'}) \eps_{v,u}^m| + |\sum_{(v,u)\in \mathcal{T}_{k,H}^{m,i}} \clip{(x_{v,u}^m)^\T\mu'} \eps_{v,u}^m| 
%    \\&\le \sum_{(v,u)\in \mathcal{T}_{k,H}^{m,i}} |\clip{(x_{v,u}^m)^\T\mu} - \clip{(x_{v,u}^m)^\T\mu'}| + |\sum_{(v,u)\in \mathcal{T}_{k,H}^{m,i}} \clip{(x_{v,u}^m)^\T\mu'} \eps_{v,u}^m| \tag{$|\eps_s|\le 1$}
%   \\&\le HK\xi_\ell  + 4\sqrt{\sum_{(v,u)\in \mathcal{T}_{k,H}^{m,i}} \clip{(x_{v,u}^m)^\T\mu'}^2 \eta_{v,u}^m \ln(1/\dt)} + 4\cd 2^{-\ell}\ln(1/\dt)
%    \\&\le HK \xi_\ell + 4\sqrt{2\sum_{(v,u)\in \mathcal{T}_{k,H}^{m,i}} \{\clip{(x_{v,u}^m)^\T\mu}^2 +\xi_\ell^2\}\eta_{v,u}^m \ln(1/\dt)} + 4\cd 2^{-\ell}\ln(1/\dt)
%    \\&\le HK \xi_\ell +4\xi_\ell\sqrt{2HK\ln(1/\dt)}+ 4\sqrt{2\sum_{(v,u)\in \mathcal{T}_{k,H}^{m,i}} \clip{(x_{v,u}^m)^\T\mu}^2 \eta_{v,u}^m \ln(1/\dt)} + 4\cd 2^{-\ell}\ln(1/\dt)
%    \\&\le 2^{-\ell} + 4\sqrt{2\sum_{(v,u)\in \mathcal{T}_{k,H}^{m,i}} \clip{{x_{v,u}^{m}}^\T\mu}^2\ln(1/\dt)} + (4\sqrt{2}+4) \cd 2^{-\ell}  \ln(1/\dt)  \tag{choose $\xi_\ell = 2^{-\ell}/(HK)$}
%    \\&\le                       4\sqrt{2\sum_{(v,u)\in \mathcal{T}_{k,H}^{m,i}}\clip{{x_{v,u}^{m}}^\T\mu}^2 \eta_{v,u}^m \ln(1/\dt)} + 12\cd 2^{-\ell}\ln(1/\dt) 
%    \tag{by $1 \le \ln(1/\dt)$ }
%  \end{align*}

  \begin{align*}
    &|\sum_{(v,u)\in \mathcal{T}_{k,H}^{m,i}}  \clip{(x_{v,u}^m)^\T \mu } \eps_{v,u}^{m}|
    \\&=   |\sum_{(v,u)\in \mathcal{T}_{k,H}^{m,i}} (\clip{(x_{v,u}^m)^\T \mu} - \clip{(x_{v,u}^m)^\T\mu'}) \eps_{v,u}^m| + |\sum_{(v,u)\in \mathcal{T}_{k,H}^{m,i}} \clip{(x_{v,u}^m)^\T\mu'} \eps_{v,u}^m| 
    \\&\le \sum_{(v,u)\in \mathcal{T}_{k,H}^{m,i}} |\clip{(x_{v,u}^m)^\T\mu} - \clip{(x_{v,u}^m)^\T\mu'}| + |\sum_{(v,u)\in \mathcal{T}_{k,H}^{m,i}} \clip{(x_{v,u}^m)^\T\mu'} \eps_{v,u}^m| \tag{$|\eps_s|\le 1$}
    \\&\le HK\xi_\ell  + 4\sqrt{\sum_{(v,u)\in \mathcal{T}_{k,H}^{m,i}} \clip{(x_{v,u}^m)^\T\mu'}^2 \VV[\eps^m_{v,u} \mid \cF_{u}^v] \ln(1/\dt)} + 4\cd 2^{-\ell}\ln(1/\dt)
    \\&\le HK \xi_\ell + 4\sqrt{2\sum_{(v,u)\in \mathcal{T}_{k,H}^{m,i}} \{\clip{(x_{v,u}^m)^\T\mu}^2 +\xi_\ell^2\}\VV[\eps^m_{v,u} \mid \cF_{u}^v] \ln(1/\dt)} + 4\cd 2^{-\ell}\ln(1/\dt)
    \\&\le HK \xi_\ell +4\xi_\ell\sqrt{2HK\ln(1/\dt)}+ 4\sqrt{2\sum_{(v,u)\in \mathcal{T}_{k,H}^{m,i}} \clip{(x_{v,u}^m)^\T\mu}^2 \VV[\eps^m_{v,u} \mid \cF_{u}^v] \ln(1/\dt)} + 4\cd 2^{-\ell}\ln(1/\dt)
    \\&\le (4\sqrt{2}+5) \cd 2^{-\ell}  \ln(1/\dt) + 4\sqrt{2\sum_{(v,u)\in \mathcal{T}_{k,H}^{m,i}} \clip{({x_{v,u}^{m}})^\T\mu}^2\VV[\eps^m_{v,u} \mid \cF_{u}^v]\ln(1/\dt)}  \tag{choose $\xi_\ell = 2^{-\ell}/(HK)$}
        \\&\le 4 \cd 2^{-\ell}  \ln(1/\dt') + 4\sqrt{\sum_{(v,u)\in \mathcal{T}_{k,H}^{m,i}} \clip{({x_{v,u}^{m}})^\T\mu}^2\eta_{v,u}^m\ln(1/\dt')}  \tag{setting $\dt = {\dt'}^{1/3}$}
    \end{align*}
We then take union bounds over $m \in \{1,2,...,L\}$, $i,\ell \in [L]$, $k\in[K]$, and $\mu' \in \hcB_\ell$, which invoke applying Lemma \ref{ineq} $(2HK)^{2(d+2)}$ times. It follows from
\begin{equation*}
    \sum_{i,\ell,k} |\hcB_\ell| = L_0 L K \sum_{\ell} (HK2^\ell)^d \le (HK)^2 (HK)^d\frac{2^{Ld}}{2^d} \le (HK2^{L})^{d+2} \le (2HK)^{2(d+2)}.
\end{equation*}
Hence, the display above holds with probability at least $1-{\dt}(1+\log_2(HK))(2HK)^{2(d+2)}\ge 1-\dt (2HK)^{2(d+3)}$. Replacing $\dt$ with $ 1/(2HK)^{2(d+3)}\cd \dt$ and setting $\iota=3\cd\ln((2HK)^{2(d+3)}/\dt)$ the result follows.

\end{proof}

%%%%%%%%%%%%%%%%%%%%%%%%%%%%%%%%%%%%%%%%%%%%%%%%%%%%%%%%%%%%%%%%%%%%%%%%%%%%%%%%%%%%%%%%%%%%%%%%

\subsection{Proof of Lemma \ref{lem:epc2}}
\begin{proof}
  Lemma 12 of~\citet{ay11improved} shows the following in its proof: Let $A$, $B$, and $C$ be positive semi-definite (PSD) matrices such that $A = B + C$. Then we have that 
  \begin{align*}
    \sup_{x\neq 0 }\fr{\|x\|^2_{A} }{\norm{x}^2_B}  \le \fr{\det(A)}{\det(B)} ~.
  \end{align*}
  Note that $V^{-1}_{\anc(t)} = V^{-1}_{t-1} + C$ for some PSD matrix $C$ (to see this, apply a series of rank-one update formula for the covariance matrix).
  Thus, we have that if $t \in J$, then
  \begin{align*}
    r < \fr{\norm{x_t}^2_{V_{\anc(t)}^{-1}}}{\norm{x_t}^{2}_{V_{t-1}^{-1}} }  
    \le \fr{|V^{-1}_{\anc(t)}|}{|V_{t-1}^{-1}|} 
    =   \fr{|V_{t-1}|}{|V_{\anc(t)}|}  ~.
  \end{align*}
  Let ${J[t]} = J \cap [t]$ with $[0] := \emptyset$.
  Let $\prev(t)$ be the time step in $J$ immediately prior to $t$: $\prev(t) := \max\{s\in \{0\} \cup J: s < t\}$ 
  Define ${W_t} = V_0 + \sum_{s\in J[t]} x_s x_s^\T$.
  \begin{align*}
    \del{\fr{d\lam + X^2|J|}{d} }^d
    &\ge \del{\fr{\tr(W_T)}{d} }^d
    \\&\ge |W_T| \tag{AM-GM ineq.}
    \\&= |V_0| \prod_{t\in J} \fr{|W_t|}{|W_{\mathsf{prev}(t)}|} 
    \\&\ge |V_0| \prod_{t\in J} \fr{|W_t|}{|W_{\anc(t)}|}  \tag{$ \mathsf{prev}(t) \le \anc(t)$ }
    \\&= |V_0| r^{|J|}
    \\&\ge \lam^d r^{|J|}
    \\  \implies 
    |J| &\le \fr{d}{\ln(r)}\ln\del{1+\fr{X^2|J|}{d\lam} } 
  \end{align*}
  Then, 
  \begin{align} \label{eq:20210425-0} 
    |J| &\le \fr{d}{\ln(r)}\ln\del{1+\fr{X^2|J|}{d\lam} } =: {A} \ln (1 + {B} |J|)
  \end{align}
  We want to solve it for $|J|$.
  Do the following:
  \begin{align}
    |J| 
    \le A \ln(1 + B |J|) 
    &= A\del{\ln\del{\fr{|J|}{2A} } + \ln\del{2A(\fr{1}{|J|} + B) } } 
    \\&\le \fr{|J|}{2} + A \ln\del{\fr{2A}{e} \del{\fr{1}{|J|} + B } } 
    \\  \implies |J| &\le 2A \ln\del{\fr{2A}{e} \del{\fr{1}{|J|} + B } }  =  \fr{2}{\ln(r)} d \ln\del{\fr{2d}{e\ln(r)}\del{\fr 1 {|J|}  + \frac {X^2} {d\lam}}  }  \label{eq:20210425-3}
  \end{align}
  
  We fix $c>0$ and consider two cases:
  \begin{itemize}
    \item Case 1: $|J| < cd$ \\
    In this case, from~\eqref{eq:20210425-0}, we have $|J| \le \fr{d}{\ln(r)} \ln\del{1 + \fr{cL^2}{\lam} }  $ 
    \item Case 2: $|J| \ge cd$\\
    In this case, from~\eqref{eq:20210425-3} we have $ |J| \le  \fr{2}{\ln(r)} d \ln\del{\fr{2}{e \ln(r)}\del{\fr 1 {c}  + \frac{X^2}{ \lam}}  }$
  \end{itemize}
  We set $c = \fr{2}{e \ln(r)}$ to obtain
  $   |J| \le \frac{2}{\ln(r)} d \ln \del{1 + \fr{2/e}{\ln(r)}\frac {X^2} {\lam}}  $.
\end{proof}

\subsection{Proof of Lemma \ref{lem:concreg-emp-rl}}
%\gray{ \begin{align*}
%    \|\mu_{k,h}^m \|_{(W_{\anc(a)}^{m,i,\ell} - 2^{-\ell} I )}^2 
%    &= \sum_{(v,u)\in \mathcal{T}_{k-1,H}^{m,i}} \clip{(x_{v,u}^m)\mu_{k,h}^m} (x_{v,u}^m)\mu_{k,h}^m 
%    \\&= \sum_{(v,u)\in \mathcal{T}_{k-1,H}^{m,i}} \clip{(x_{v,u}^m)\mu_{k,h}^m} (-\eps_{v,u}^m(\th_{k,h}^m) + \eps_{v,u}^m(\th^*)) 
%    \\&\le  \sqrt{\sum_{(v,u)\in \mathcal{T}_{k-1,H}^{m,i}} \clip{x_{v,u}\mu_{k,h}^m}^2 \eta_{v,u}^m(\th_{k,h}^m) \iota}  + 4\cd 2^{-\ell} \iota + 4\sqrt{\sum_{(v,u)\in \mathcal{T}_{k-1,H}^{m,i}} \clip{(x_{v,u}^m)\mu_{k,h}^m}^2 \eta_{v,u}^m(\th^*) \iota} + 4\cd2^{-\ell} \iota 
%    \\&\le 2\sqrt{\sum_{(v,u)\in \mathcal{T}_{k-1,H}^{m,i}} \clip{(x_{v,u}^m)\mu_{k,h}^m}^2\cd 2^{-i} \iota}+8\cd 2^{-\ell}\iota 
%        \\&\le 2\sqrt{\sum_{(v,u)\in \mathcal{T}_{k-1,H}^{m,i}} \clip{(x_{v,u}^m)\mu_{k,h}^m}x_{v,u}^m)\mu_{k,h}^m \cd 2^{-i} \iota}+8\cd 2^{-\ell}\iota 
%    \\& \le 2\|\mu_{k,h}^m\|_{W_{\anc(a)}^{m,i,\ell}}\sqrt{2^{-i}\iota}+8\cd 2^{-\ell}\iota
%  \end{align*}
%Solving for $\|\mu_{k,h}^m\|_{W_{\anc(a)}^{m,i,\ell}}$, we get the desired bound.
%}
%\kj{how about the following?}

Recall that $\th_t^m=\argmax_{\th \in \Th_{k-1}} |\{\th x_t^{m+1}-(\th x_t^{m})^2\}|$ and $\eta_b^m(\th)=\th x_b^{m+1}-(\th x_b^{m+1})^2$
  We have
  \begin{align*}
  \|\mu_{t}^m \|_{(W_{\anc(t)}^{m,i,\ell} - 2^{-\ell} I )}^2 
  &= \sum_{b\in \mathcal{T}_{\anc(t)}^{m,i}} \clip{(x_{b}^m)^\top\mu_{t}^m} (x_{b}^m)\mu_{t}^m 
  \\&= \sum_{b\in \mathcal{T}_{\anc(t)}^{m,i}} \clip{(x_{b}^m)^\top\mu_{t}^m} (-\eps_{b}^m(\th_{t}^m) + \eps_{b}^m(\th^*)) 
  \\&\le  \sqrt{\sum_{b\in \mathcal{T}_{\anc(t)}^{m,i}} \clip{(x_{b}^m)^\top\mu_{t}^m}^2 \eta_{b}^m(\th_{t}^m) \iota}  + 4\cd 2^{-\ell} \iota + 4\sqrt{\sum_{b\in \mathcal{T}_{\anc(t)}^{m,i}} \clip{(x_{b}^m)^\top\mu_{t}^m}^2 \eta_{b}^m(\th^*) \iota} + 4\cd2^{-\ell} \iota 
  \\&\le C\sqrt{\sum_{b\in \mathcal{T}_{\anc(t)}^{m,i}} \clip{(x_{b}^m)^\top\mu_{t}^m}^2\cd 2^{-i} \iota}+8\cd 2^{-\ell}\iota %\tag{\kj{$\eta^m_b \le 2\cd 2^{-i}$, and $\eta^m_b(\th^*) \le \eta^m_b$, right? do we have the right const here?}}
  \\&\le C\sqrt{\sum_{b\in \mathcal{T}_{\anc(t)}^{m,i}} \clip{(x_{b}^m)^\top\mu_{t}^m} (x_{b}^m)^\T\mu_{t}^m \cd 2^{-i} \iota}+8\cd 2^{-\ell}\iota
  \\& \le C\|\mu_{t}^m\|_{W_{\anc(t)}^{m,i,\ell}}\sqrt{2^{-i}\iota}+8\cd 2^{-\ell}\iota
\end{align*}
Solving for $\|\mu_{t}^m\|_{W_{\anc(t)}^{m,i,\ell}}$, we get the desired bound.

\subsection{The Exact Definition of $I_{k,h}$ and Proof of Lemma~\ref{lem-I}}\label{proof_of_lem_I}

%Fix $m$, $i$, $\ell$ and let us denote $x_{k,h}$ and $\mu_{k,h}$ by $x_a$ and $\mu_a$ with $D[a]=(k,h)$. %Hence we can further define $\cT_a^{m,i}:=\cT_{D[a]}^{m,i}$. 
Fix $m$ and $i$.
Let $t \in \cT^{m,i}_T$ and $\ell = \ell^m_t$.
We recall Lemma~\ref{lem:concreg-emp-rl} that yields
\begin{align*}
 \norm{\mu^m_{t}}_{W^{m,i,\ell}_{\anc(t)}} 
  &\le c_M \cd \max\{\sqrt{2^{-i} \iota}, \sqrt{2^{-\ell}\iota} \}
\end{align*}

Let us define $\blue{\mathcal{T}^{m,i,\ell}}=\{t \in \cT_T^{m,i}: (x_{t}^m)^\T \mu_{t}^m \in (2\cd2^{-\ell},2\cd 2^{1-\ell}])\}$
and split the time steps $\cT^{m,i,\ell}$  as follows:
\begin{align*}
  \blue{\cT^{m,i,\ell,\la1\ra}} := \cbr{t \in \cT^{m,i,\ell}: \norm{\mu^m_{t}}_{W^{m,i,\ell}_{\anc(t)}} \le c_M \sqrt{2^{-i} \iota} }  
  \text{~~~ and ~~~} 
  \blue{\cT^{m,i,\ell,\la2\ra}} := \cT^{m,i,\ell}  \sm \cT^{m,i,\ell,\la1\ra} ~.
\end{align*}

Given $t \in [T]$, let $\blue{n_t^{m,i,\ell,\la1\ra}}$ be $n \in \{0,\ldots,\ell\} $ such that
\begin{align*}
  \max_{t':t \le t'\in \cT^{m,i,\ell,\la1\ra}} |(x^m_{t})^\T\mu^m_{t'}| \in \lt\lparen \onec{n \neq 0} \cd 2^{-\ell+n}, 2^{-\ell+n+1}  \rt\rbrack 
\end{align*}
where we set $n=0$ if the maximum above is less or equal to $2^{-\ell}$.

Let $\blue{\cT^{m,i,\ell,\la1\ra,n}} = \cT^{m,i,\ell,\la1\ra} \cap \{t: n_t^{m,i,\ell,\la1\ra} = n\}$ and
set $\blue{V^{m,i,\ell,\la1\ra,n}_{t}} := 2^{-\ell}I + \sum_{b \in \cT^{m,i,\ell,\la1\ra,n}_{t}} x^m_{b}  (x^m_{b})^\T$ with $\blue{\cT_t^{m,i,\ell,\la z \ra,n}} :=\{a'\le a:a'\in \cT^{m,i,\ell,\la z \ra,n}\}$.

Define
\begin{align*}
    \blue{I^{m,i,\ell,\la1\ra,n}_{t}} := \onec{\forall t'\in \cT^{m,i,\ell,\la1\ra,n}_{t} \text{ with } \sfk(t')=\sfk(t),~   \norm{x_{t'}}^2_{(V^{m,i,\ell,\la1\ra,n}_{\anc(t')})^{-1}} \le r \norm{x_{t'}}^2_{(V^{m,i,\ell,\la1\ra,n}_{t'-1})^{-1}}  }
\end{align*}
for some $r>1$ to be specified later.
With this definition, one can see that $\sum_{t\in[T]: \sfh(t) < H}  I^{m,i,\ell,\la1\ra,n}_{t}-I^{m,i,\ell,\la1\ra,n}_{t+1}$ is the number of bad episodes where there exists $h\in[H]$ such that 
\begin{align*}
 \norm{x_{t}}^2_{(V^{m,i,\ell,\la1\ra,n}_{\anc(t)})^{-1}} > r \norm{x_t}^2_{(V^{m,i,\ell,\la1\ra,n}_{t-1})^{-1}}~.
\end{align*}
Define $\blue{I^{m,i,\ell,\la1\ra,n}_{k,h}} := I^{m,i,\ell,\la1\ra,n}_{\sft(k,h)}$.
%\kj{how about $\sfk(t') = k-1$ instead of $D[a][1]=k$?}
Define similar quantities for $\la2\ra$ as well.

Define $\blue{I_{t}} := \prod_{m,i,\ell,n} I^{m,i,\ell,\la1\ra,n}_{t} \prod_{m,i,\ell,n} I^{m,i,\ell,\la2\ra,n}_{t}$.
We now prove Lemma~\ref{lem-I}.

Note that
\begin{align*}
\sum_{k=1}^K \sum_{h=1}^{H-1} I_{k,h} - I_{k,h+1}  \le \sum_{k=1}^K \sum_{h=1}^{H-1} \sum_{m \le L_0}\sum_{i \le L }\sum_{\ell \le L} \sum_{z\in[2]} \sum_{n =0}^{\ell} \del{I^{m,i,\ell,\la z\ra,n}_{k,h} - I^{m,i,\ell,\la z\ra,n}_{k,h+1} },
\end{align*}
We now assume $z=1$ without loss of generality.
The display above can be written as
\begin{align*}
    \sum_{t\in[T]: \sfh(t) < H}  (I_{t} - I_{t+1})
  &\le \sum_{m,i,\ell,n} \sum_{t\in[T]: \sfh(t) < H}  I^{m,i,\ell,\la1\ra,n}_{t} - I^{m,i,\ell,\la1\ra,n}_{t+1}~.
\end{align*}
% Recalling that $\sum_{t\in[T]: \sfh(t) < H}  I^{m,i,\ell,\la1\ra,n}_{t}-I^{m,i,\ell,\la1\ra,n}_{t+1}$ is counting the number of bad episodes $k$ where there exists $h\in[H]$ such that 
% \begin{align*}
%  \norm{x_{t}}^2_{(V^{m,i,\ell,\la1\ra,n}_{\anc(t)})^{-1}} > r \norm{x_t}^2_{(V^{m,i,\ell,\la1\ra,n}_{t-1})^{-1}}
% \end{align*}
Recall that the inner sum above is the count of the `bad' episodes.
We invoke Lemma~\ref{lem:epc2} $2L_0(L)^3$ times with $X=\sqrt{d}$ as $\|x_t\|_1\leq \sqrt{d}$  to finish the proof as 
 \begin{equation*}
    \sum_{k,h}I_{k,h}-I_{k,h+1}\le O\left(\frac{d}{\ln (r)}\log^5\left(dHK(1 + \fr{d^2}{\ln (r) }) \right)\right)
\end{equation*}
%\kj{I changed the above slightly; does it look correct?}
%{\color{red} You are right, I think this can be simplified since
%$$C\log(dHK) \ge \log(dHK(1+\frac{1}{\ln(r)}))$$
%for $C>r$}

\subsection{Proof of Proposition ~\ref{prop:rl-bound}}\label{details}

To proceed we safely choose $r=2$ and inherit all notations from Section~\ref{proof_of_lem_I}. Let us define
\begin{align*}
  \blue{\brcT^{m,i,\ell,\la z\ra}} := \cT^{m,i,\ell,\la z\ra} \cap \cbr{t\in[T]:I_t = 1}  
\end{align*}
where $z \in \{1,2\}$.
We start with $R_m$ as 
\begin{align*}
  R_m &= \sum_{t}  (\brx_{t}^m)^\top \mu_{t}^m
  \\&\le \sum_{t,i,\ell} \sum_{t \in \brcT^{m,i,\ell,\la1\ra}}  (\brx_{t}^m)^\top \mu_{t}^m
  + \sum_{t,i,\ell} \sum_{t \in \brcT^{m,i,\ell,\la2\ra}} (\brx_{t}^m)^\top \mu_{t}^m.
\end{align*}

Let us fix $m$, $i$, and $\ell$ and focus on controlling $\sum_{t \in \brcT^{m,i,\ell,\la z\ra}} (\brx_{t}^m)^\top \mu_{t}^m $ for $z\in[2]$.
Hereafter, we omit the superscripts of $(m,i,\ell)$ to avoid clutter, unless there is a need. 

Note that for $t  \in \brcT^{\la1\ra,n}$ and $b$ such that $t \le b$, 
\begin{align*}
  W_{\anc(t)}(\mu_b)
  &= 2^{-\ell} I +  \sum_{t' \in \cT^{m,i,\ell}_{\anc(t)}} (1\wedge \fr{2^{-\ell} }{|x_{t'} \mu_b | } )   x_{t'} x_{t'}^\T
  \\&\succeq 2^{-\ell} I +  \sum_{t' \in \cT^{m,i,\ell,\la1\ra,n}_{\anc(t)}  } (1\wedge \fr{2^{-\ell} }{|x_{t'} \mu_b | } )   x_{t'} x_{t'}^\T
  \\&\succeq 2^{-\ell} I +  \sum_{t' \in \cT^{m,i,\ell,\la1\ra,n}_{\anc(t)}  } (1\wedge \fr{2^{-\ell} }{2^{-\ell+n+1} } )   x_{t'} x_{t'}^\T
  \\&\succeq 2^{-\ell} I +  2^{-n-1} \sum_{t' \in \cT^{m,i,\ell,\la1\ra,n}_{\anc(t)} } x_{t'} x_{t'}^\T
  \\&\succeq c2^{-n} V^{\la1\ra,n}_{\anc(t)}~.
\end{align*}

For $t  \in \brcT^{\la1\ra,n}$, let $b = \arg \max_{t \le b'\in \brcT^{\la1\ra} } |x_{t}^\top \mu_{b'}|$.
%\kj{inner product notation}
Then,
\begin{align*}
  2^{-\ell + n} 
  &\le |x_{t}^\top \mu_{b}| \tag{def'n of $b$}
  \\&\le \norm{x_{t}}_{W_{\anc(t)}^{-1}(\mu_{b})}  \norm{\mu_b}_{W_{\anc(t)}(\mu_b)}  
  \\&\le \sqrt{2^n} \norm{x_{t}}_{(V^{\la1\ra,n}_{\anc(t)})^{-1}}  \norm{\mu_{b}}_{W_{\anc(b)}(\mu_b)}   
  \\&\le c\sqrt{r} \sqrt{2^n} \norm{x_{t}}_{(V^{\la1\ra,n}_{t-1})^{-1}}  \sqrt{2^{-i}\iota} ~.  \tag{by $b \in \brcT^{\la1\ra}$  }
\end{align*}
This implies that
\begin{align*}
  \norm{x_{t}}_{(V^{\la1\ra,n}_{t-1})^{-1}}^2  \ge c_{\la1\ra} \fr{2^{-2\ell+n}}{r 2^{-i}\iota}. 
\end{align*}
for some absolute constant $c_{\la1\ra} > 0$.

Thus,
\begin{align*}
  \sum_{t\in \brcT^{\la1\ra}} x_{t}\mu_{t}
  &\le   c2^{-\ell} \sum_{t\in \brcT^{\la1\ra}} 1
  \\&\le c2^{-\ell} \sqrt{|\brcT^{\la1\ra}| \sum_{t\in \brcT^{\la1\ra}} 1}
  \\&\le c2^{-\ell} \sqrt{|\brcT^{\la1\ra}| \sum_{n=0}^{\ell} \sum_{t\in \brcT^{\la1\ra,n}} 1}
  \\&\le c2^{-\ell} \sqrt{|\brcT^{\la1\ra}| \sum_{n=0}^{\ell} \sum_{t\in \brcT^{\la1\ra,n}} \onec{\norm{x_t }_{(V^{\la1\ra,n}_{t-1})^{-1}}^2 \ge c_{\la1\ra}\fr{2^{-2\ell+n}}{r 2^{-i}\iota }} } 
  \\&\le c2^{-\ell} \sqrt{|\brcT^{\la1\ra}| \sum_{n=0}^{\ell} \sum_{t\in \cT^{\la1\ra,n}} \onec{\norm{x_t }_{(V^{\la1\ra,n}_{t-1})^{-1}}^2 \ge c_{\la1\ra}\fr{2^{-2\ell+n}}{r 2^{-i}\iota }} } \tag{$\brcT^{\la1\ra,n} \subseteq \cT^{\la1\ra,n}$}
  \\&\le c2^{-\ell} \sqrt{|\brcT^{\la1\ra}| \sum_{n=0}^{\ell} \fr{r 2^{-i} \iota}{2^{-2\ell + n} } d \ln\del{ 1 + c\fr{r 2^{-i}\iota}{2^{-2\ell+n} 2^{-\ell} }  }  }  \tag{Lemma~\ref{lem:epc}}
  \\&\le c\sqrt{dr |\brcT^{\la1\ra}| 2^{-i} \iota \ln\del{ 1 + c r \iota 8^{\ell}  }  } 
  \\&\le c \sqrt{dr (1 + \sum_{t\in \brcT^{\la1\ra}} \eta_t ) \iota \ln\del{ 1 + c r \iota 8^{\ell}  }  } 
\end{align*}
% where $\blue{\breve\eta^m_t} := \eta^m_t I^{m}_t$ with $I^m_t := \prod_i \prod_\ell \prod_n \prod_{z\in\{1,2\}} I^{m,i,\ell,\la z\ra,n}_t$.
%$\bar\eta = \sum_{t \in \brcT^{\la1\ra}} \breve\eta^m_{t} $. 

For the other case involving $\brcT^{\la2\ra}$, we use the same logic as above.
Let $t \in \brcT^{\la1\ra,n}$.
Then, for any $b$ such that $a \le b$, one can show that
\begin{align*}
  W_{\anc(t)}(\mu_b)
  &\succeq c2^{-n} V^{\la2\ra,n}_{\anc(t)}~.
\end{align*}
Then, once again letting $b = \arg \max_{t<b\in \brcT^{\la2\ra} } |x_{t}^\top \mu_{b}|$ for $t\in\brcT^{\la2\ra,n}$ one can show that
\begin{align*}
  2^{-\ell + n} 
  &\le c\sqrt{r} \sqrt{2^n} \norm{x_{t}}_{(V^{\la2\ra,n}_{t-1})^{-1}}  \sqrt{2^{-\ell}\iota} ~.
\end{align*}
This implies that
\begin{align*}
  \norm{x_t}_{(V^{\la2\ra,n}_{t-1})^{-1}}^2  \ge c \fr{2^{-\ell+n}}{r \iota} 
\end{align*}
for some absolute constant $c>0$.
Then,
\begin{align*}
  \sum_{t\in \brcT_{\la2\ra}} x_t^\top\mu_t
  &\le   c2^{-\ell} \sum_{n=0}^{\ell} \sum_{(k,h)\in \brcT^{\la2\ra,n}} 1 
  \\&\le   c2^{-\ell} \sum_{n=0}^{\ell} \sum_{t\in \brcT^{\la2\ra,n}} \onec{\norm{x}^{2}_{(V^{\la2\ra,n}_{a-1})^{-1}} \ge c\fr{2^{n-\ell}}{r\iota} }
  \\&\le   c2^{-\ell} \sum_{n=0}^{\ell} c \fr{r\iota}{2^{n-\ell} }   d\ln\del{1+c\fr{r\iota}{2^{n-\ell}\cd 2^{-\ell} } }
  \\&\le   c d r \iota \ln\del{1 + c r \iota 4^\ell}
\end{align*}

Invoking the elliptical potential count lemma together with $r=2$,
\begin{align*}
  R_m \le c\sum_{i}^{L} \sum_{\ell}^{L}   \del{d^{0.5}\sqrt{(1+ \sum_{t\in\brcT^{m,i,\ell}}\eta^m_{t})\iota \ln\del{1+c\iota 8^\ell} } + d \iota\ln\del{1+c\iota 8^\ell} },
\end{align*}
which implies that
\begin{align*}
  R_m &\le  O\left(d^{0.5}\log^{2.5}(HK)\sqrt{\sum_{t} \breve\eta^m_t \iota \log(\iota )}+d \log^3(HK)\iota \log(\iota)\right).
  %\\&\le \hat O(d^{2.5}\log^{3}(dHK)\sqrt{\bar \eta \iota  )}+d^6 \log^{3.5}(dHK)\iota)
\end{align*}
where $\blue{\breve\eta^m_t} := \eta^m_t I_t$.

\subsection{Proof of Theorem \ref{main-rl}}\label{proof_of_rl}

\begin{proof}
We continue from the proof in the main paper where it remains to bound $R_0 + M_0$.
Using the relation  (equation (56) and (57) in \citep{zhang21variance}),
\begin{equation*}
    \sum_{t}\breve \eta_{t}^m \le |M_{m+1}|+O(d\log^5(dHK)+2^{m+1}(K+R_0))+R_{m+1}+2R_m,
\end{equation*}
one has, using Proposition~\ref{prop:rl-bound}, 
\begin{equation*} 
    R_m\le O\big(d^{0.5}   \log^{2.5}(dHK)\sqrt{\iota\log(\iota)}\sqrt{|M_{m+1}|+2^{m+1}(K+R_0)+R_{m+1}+2R_m}+d \log^{5}(dHK) \iota\log(\iota)) 
\end{equation*}
The strategy is to solve the recursive inequalities with respect to $R_m$ and $M_m$ to obtain a bound on $R_0$ and $M_0$.
By Lemma~\ref{lem-M}, we have
\begin{equation}\label{boot2}
|M_m| \le O(\sqrt{|M_{m+1}| + 2^{m+1}(K+R_0)\log(1/\delta)}+d^{0.5}\log^{2.5}(dHK) + \log(1/\delta))~.
\end{equation}
With $\blue{b_m} :=R_m+|M_m|$, we have
\begin{equation*}
    b_m 
    \le 
    \hat O\big(d \log^{3}(dHK) \sqrt{\log(1/\dt)}\sqrt{b_m+b_{m+1}+2^{m+1}(K+R_0)} + d^{2} \log^{7}(dHK)\log (1/\dt) \big).
\end{equation*}
where $\hat O$ ignores doubly logarithmic factors. 

We now use Lemma~\ref{rec-lem} with $\lambda_1=HK$, $\lambda_2= \hat\Theta(d\log^{3}(dHK)\sqrt{\log (1/\dt)})$, $\lambda_3=(K+R_0)$ and $\lambda_4=\hat\Theta(d^{2} \log^{7}(dHK)\ln(1/\dt))$ where $\hat\Theta$ ignores doubly logarithmic factors, we obtain
\begin{equation*}
    R_0\le b_0 \le \tilde O\big(d^{2}\log^{7}(dHK)\log (1/\dt)+\sqrt{(K+R_0)d^2\log^6(dHK)\log(1/\dt)}\big).
\end{equation*}
We can solve it for $R_0$ to obtain $R_0\le \tilde O(\sqrt{K d^2\log^7(dHK)\log(1/\dt)} + d^{2}\log^{7}(dHK)\log (1/\dt))$. 

Next, we apply Lemma~\ref{bootstrap} to \eqref{boot2} with $\lambda_2=\Theta(1)$, $\lambda_3 = (K+R_0)\ln(1/\dt)$, and $\lambda_4 = \Theta(d^{0.5} \log^{2.5}(dHK) + \ln(1/\dt))$ to obtain
\begin{align*}
    |M_0| 
    &\le O(\sqrt{(K+R_0)\ln(1/\dt)} + d^{0.5} \log^{2.5}(dHK) + \ln(1/\dt))
  \\&\le \tilde O(\sqrt{K\log(1/\dt)} + \sqrt{d^{2}\log^{7}(dHK)\log^2(1/\dt)} + d^{0.5} \log^{2.5}(dHK) + \ln(1/\dt))
\end{align*}
where the last inequality uses $\sqrt{AB} \le \frac{A+B}{2}$ to obtain the following:
\begin{align*}
    K + R_0 
      &\le \tilde O( K + d^{2}\log^{7}(dHK)\log (1/\dt)+\sqrt{K d^2\log^7(dHK)\log(1/\dt)} )
    \\&\le \tilde O\del{ K + d^{2}\log^{7}(dHK)\log (1/\dt) + \fr12\cd K + \fr12\cd d^{2}\log^{7}(dHK)\log (1/\dt) }~.
\end{align*}
Altogether, we obtain
\begin{align*}
    b_0 = \tilde O( \sqrt{K d^2\log^7(dHK)\log^2(1/\dt)} + d^{2}\log^{7}(dHK)\log (1/\dt) )
\end{align*}
This concludes the proof.

\end{proof}

\subsection{Miscellaneous lemmas}

\begin{lemma}\label{ineq}(\citet[Lemma 11]{zhang2021model})
  Let $(M_n)_{n\geq0}$ be a martingale such that $M_0=0$ and $|M_n-M_{n-1}| \leq b$ almost surely for $n\geq1$. For each $n\geq 0$, let $\mathcal{F}_n=\sigma(M_1,...,M_n)$. Then for any $n \geq 1$ and $\eps,\delta>0$, we have
  \begin{equation*}
      \mathbb{P}\del{|M_n|\geq 2 \sqrt{ \sum_{i=1}^n \mathbb{E}[(M_{i}-M_{i-1})^2|\mathcal{F}_{i-1}] \ln(1/\delta)}+2\sqrt{\eps \ln(1/\delta)}+2 b \ln(1/\delta)} \leq 2(\log_2(b^2 n/\eps) +1)\delta
  \end{equation*}
\end{lemma}

\begin{lemma}\label{lem-M}(\citet[Lemma 25]{zhang21variance})
\begin{align*}
  |M_m| \le O\del{\sqrt{M_{m+1}+O(d\log^5(dHK))+2^{m+1}(K+R_0)\log(1/\delta)}+\log(1/\delta)}
\end{align*}
\end{lemma}

%\begin{proof}
%It is a straightforward result by modifying $d\log^5(dHK)$ in Lemma 12 \citep{zhang21variance} to $d\log^4(dHK)$ using Lemma~\ref{lem-I}
%\end{proof}
\begin{lemma}\label{lem-R3}(\citet[Lemma 6]{zhang21variance})
$\Reg_3\le O(\sqrt{K\log(1/\dt)})$.
\end{lemma}

\begin{lemma}\label{rec-lem}(\citet[Lemma 12]{zhang21variance})
For $\lambda_i>0$, $i\in\{1,2,4\}$ and $\lambda_3 \ge 1$, let $\kappa=\max\{\log_2(\lambda_1),1\}$. Assume that $0\le a_i\le \lambda_1$ and $a_i \le \lambda_2 \sqrt{a_i+a_{i+1}+2^{i+1}\lambda_3}+\lambda_4$ for $i \in \{1,2,...,\kappa\}$. 
Then, we have
\begin{equation*}
    a_1 \le 22 \lambda_2^2 +6\lambda_4+4\lambda_2 \sqrt{2\lambda_3}
\end{equation*}
\end{lemma}

\begin{lemma}\label{bootstrap}(\citet[Lemma 2]{zhang2021reinforcement})
Let $\lambda_1,\lambda_2,\lambda_4 \ge0$ and $\lambda_3 \ge 1$ with $i'=\log_2(\lambda_1)$. We have a sequence $\{a_i\}_i$ for $i \in \{1,2,...,i'\}$ satisfying $a_i\le\lambda_1$ and $a_i \le \lambda_2 \sqrt{a_{i+1}+2^{i+1}\lambda_3}+\lambda_4$. Then,
\begin{equation*}
    a_1 \le \max\cbr{(\lambda_2 +\sqrt{\lambda_2^2+\lambda_4})^2,\lambda_2\sqrt{8\lambda_3}+\lambda_4}
\end{equation*}
\end{lemma}

\section{Comparison with~\citet{he2021logarithmic}}\label{app:diff}

At a high-level, \citet{he2021logarithmic} apply a similar strategy to ours. One immediate difference is that we do not incur an extra dependence on $d$ inside the logarithm, but it could be due to the fact that their lemma is applying the count on a different quantity from ours. Note that our EPC is still novel and tighter than the bound appeared in a concurrent work of \citet[Lemma 6.2]{wagenmaker2022first} for a large enough $K$ (i.e., time horizon). 

Note that the core of our novelty is the point of view introduced by the the matrix norm with respect to $W_{\ell,k-1}(\mu)$. 
This enables the connection to the elliptical potential lemma and the peeling technique. 
Such a viewpoint is exactly what the paper of VOFUL~\cite{zhang21variance} did not seem to have realized.
Indeed, the proof of VOFUL \cite{zhang21variance} does not use peeling on $x_k^\top \mu_k$ as we do.

% Indeed, the proof of VOFUL \cite{zhang21variance} does not use peeling on $x_k^\top \mu_k$ as we do!

% , which has made the analysis rather complicated. 
% This
% Indeed, the proof of VOFUL \cite{zhang21variance} does not use peeling on $x_k^\top \mu_k$. 

We also like to highlight that Lemma~\ref{lem:epc2} is still novel and is one of the main contributors to the improved regret bound for the linear mixture MDP.